\documentclass[lettersize,journal]{IEEEtran}
\usepackage{amsmath,amsfonts}
\usepackage{algorithmicx}
\usepackage{algorithm}
\usepackage{algpseudocode}
\usepackage{array}
\usepackage[caption=false,font=normalsize,labelfont=sf,textfont=sf]{subfig}
\usepackage{textcomp}
\usepackage{stfloats}
\usepackage{url}
\usepackage{verbatim}
\usepackage{graphicx}
\usepackage{cite}
\newcommand{\revs}[1]{\textcolor{black}{#1}}

\usepackage{amsmath,amsfonts,bm}

\def\eqref#1{equation~\ref{#1}}

\def\1{\bm{1}}

\def\vv{{\bm{v}}}

\def\vx{{\bm{x}}}

\def\mD{{\bm{D}}}

\def\mH{{\bm{H}}}

\def\mP{{\bm{P}}}

\def\mS{{\bm{S}}}

\def\mW{{\bm{W}}}
\def\mX{{\bm{X}}}
\def\mY{{\bm{Y}}}
\def\mZ{{\bm{Z}}}

\DeclareMathAlphabet{\mathsfit}{\encodingdefault}{\sfdefault}{m}{sl}
\SetMathAlphabet{\mathsfit}{bold}{\encodingdefault}{\sfdefault}{bx}{n}

\def\gD{{\mathcal{D}}}
\def\gE{{\mathcal{E}}}

\def\gG{{\mathcal{G}}}

\def\gK{{\mathcal{K}}}
\def\gL{{\mathcal{L}}}

\def\gN{{\mathcal{N}}}

\def\gT{{\mathcal{T}}}

\def\gV{{\mathcal{V}}}

\def\gY{{\mathcal{Y}}}

\def\sR{{\mathbb{R}}}

\DeclareMathOperator*{\argmin}{arg\,min}

\usepackage{tabularx}
\newcolumntype{R}{>{\raggedleft\arraybackslash}X}
\newcolumntype{C}{>{\centering\arraybackslash}X}
\usepackage{threeparttable}
\usepackage{booktabs}
\usepackage{cleveref}
\usepackage{multicol}
\usepackage{multirow}

\crefname{equation}{Eq.}{Eqs.}          
\Crefname{equation}{Equation}{Equations} 

\crefname{algorithm}{Alg.}{Algs.}          
\Crefname{algorithm}{Algorithm}{Algorithms} 

\crefname{figure}{Fig.}{Figs.}          
\Crefname{figure}{Figure}{Figures}      

\crefname{table}{Tab.}{Tabs.}           
\Crefname{table}{Table}{Tables}         

\crefname{section}{Sec.}{Secs.}         
\Crefname{section}{Section}{Sections}   

\usepackage{xcolor}

\begin{document}

\title{Hypergraph Foundation Model}

\author{
Yue~Gao,~\IEEEmembership{Senior~Member,~IEEE,}
Yifan~Feng,
Shiquan~Liu,
Xiangmin~Han,~\IEEEmembership{Member,~IEEE,}\\
Shaoyi~Du,~\IEEEmembership{Member,~IEEE,}
Zongze~Wu*,~\IEEEmembership{Member,~IEEE,}
Han~Hu*,~\IEEEmembership{Member,~IEEE}
\thanks{
Yue Gao, Yifan Feng, and Xiangmin Han are with the School of Software, BNRist, THUIBCS, BLBCI, Tsinghua University, Beijing 100084, China. 
E-mail: kevin.gaoy@gmail.com; evanfeng97@gmail.com; simon.xmhan@gmail.com; }
\thanks{Shiquan Liu, and Shaoyi Du are with Institute of Artificial Intelligence and Robotics, College of Artificial Intelligence, Xi’an Jiaotong University, Xi’an 710049, China. 
E-mail: quan3759@stu.xjtu.edu.cn; 
dushaoyi@xjtu.edu.cn;}
\thanks{Zongze Wu is with the College of Mechatronics and Control Engineering, Shenzhen University, Shenzhen, Guangdong 510006, China.
E-mail: zzwu@szu.edu.cn.}
\thanks{Han Hu is with the Beijing Institute of
 Technology, Beijing 100811, China. 
E-mail: hhu@bit.edu.cn. \protect\\ 
(Corresponding authors: Zongze Wu and Han Hu)
}
}


\markboth{IEEE TRANSACTIONS ON PATTERN ANALYSIS AND MACHINE INTELLIGENCE}%
{Shell \MakeLowercase{\textit{et al.}}: A Sample Article Using IEEEtran.cls for IEEE Journals}


\maketitle

\begin{abstract}
Hypergraph neural networks (HGNNs) effectively model complex high-order relationships in domains like protein interactions and social networks by connecting multiple vertices through hyperedges, enhancing modeling capabilities, and reducing information loss. Developing foundation models for hypergraphs is challenging due to their distinct data, which includes both vertex features and intricate structural information. We present Hyper-FM, a Hypergraph Foundation Model for multi-domain knowledge extraction, featuring Hierarchical High-Order Neighbor Guided Vertex Knowledge Embedding for vertex feature representation and Hierarchical Multi-Hypergraph Guided Structural Knowledge Extraction for structural information. Additionally, we curate 11 text-attributed hypergraph datasets to advance research between HGNNs and LLMs. Experiments on these datasets show that Hyper-FM outperforms baseline methods by approximately 13.4\%, validating our approach. Furthermore, we propose the first scaling law for hypergraph foundation models, demonstrating that increasing domain diversity significantly enhances performance, unlike merely augmenting vertex and hyperedge counts. This underscores the critical role of domain diversity in scaling hypergraph models.
\end{abstract}


\section{Introduction}

Hypergraph neural networks (HGNNs) have garnered significant attention due to their ability to perform high-order association and collaborative learning on data. They have been successfully applied across various domains, including protein interaction network analysis\cite{murgas2022hypergraph,jiang2023explainable}, brain network analysis \cite{wang2023dynamic,han2024inter}, and social network analysis \cite{bick2023higher,lotito2024hyperlink}. Compared to traditional graphs, hypergraphs allow hyperedges to connect more than two vertices, thereby offering stronger modeling capabilities and reducing information loss from real-world relational data. This enhanced structural representation grants HGNNs superior data representation and learning abilities.

Foundation models have been pivotal in advancing artificial intelligence, particularly in fields like computer vision (CV) \cite{hjelm2018learning,chen2020simple} and natural language processing (NLP) \cite{fang2020cert,gao2021simcse}. In CV, models such as Vision Transformers \cite{visiontransformers} leverage large-scale datasets to learn rich visual representations, enabling superior performance across a wide range of tasks. Similarly, in NLP, models like BERT \cite{bert,roberta} and GPT \cite{gpt2,gpt3} utilize extensive textual corpora to capture intricate language patterns, facilitating advancements in language understanding and generation. However, hypergraphs represent a distinct form of high-order relational structure, with data representations fundamentally different from images and text. The input to hypergraph models comprises two distinct types of information: vertex feature representations and hypergraph structural information. This dual nature poses substantial challenges in designing and implementing foundation models for hypergraphs. Constructing a unified hypergraph foundation model for multi-domain hypergraph data necessitates bridging the diverse vertex representations and hypergraph structures across different domains. This paper explores methods to extract unified hypergraph knowledge to generate a robust hypergraph foundation model, thereby enhancing performance in target hypergraph domains. We address these challenges from two perspectives: vertex knowledge extraction and hypergraph structural knowledge extraction.

\begin{figure}[!t]
    \centering
    \includegraphics[width=\linewidth]{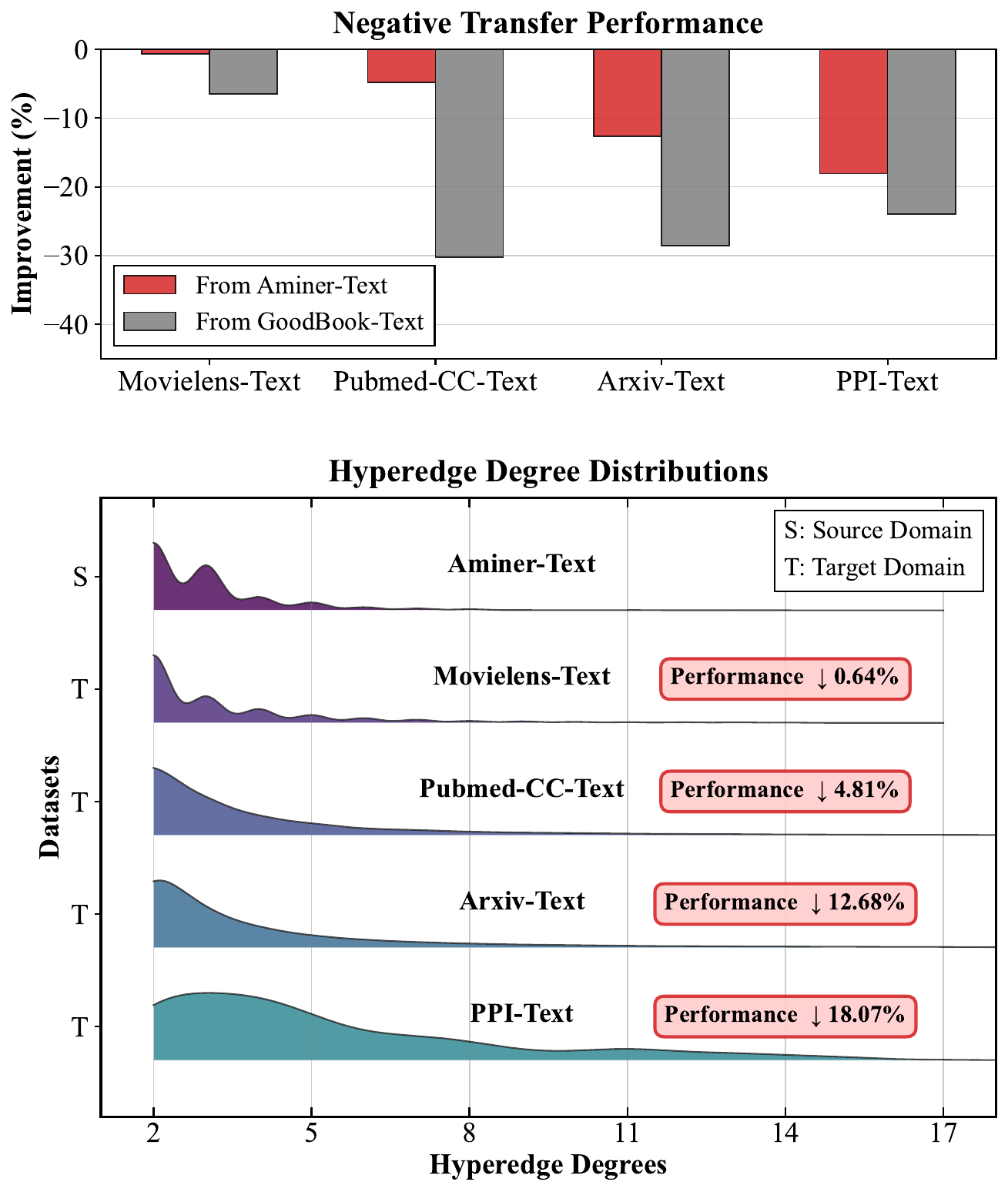}
    \caption{Illustration of negative transfer phenomenon in hypergraph datasets.}
    \label{fig:neg-trans}
\end{figure}

Regarding the extraction of knowledge from vertex features, existing hypergraph models predominantly utilize Bag-of-Words (BoW) \cite{BoW} representations. This approach leads to ever-increasing feature dimensions as textual descriptions grow, and varying vertex feature dimensions across different domains complicate the alignment of vertex features. Current methods attempt to extract fixed-dimensional embeddings from raw textual data using techniques like Singular Value Decomposition (SVD) \cite{svd} or Language Models (LM) \cite{bert,roberta,gpt2,gpt3}. However, these approaches often overlook structural information, resulting in suboptimal feature representations. For instance, vertices with identical feature representations but located in different positions within the hypergraph should possess distinct semantic embeddings. Directly using LMs for feature extraction fails to incorporate structural information, raising the question: how can structural information be injected into LM-extracted features to enhance their discriminative power?


In terms of extracting structural knowledge from hypergraphs, the most straightforward approach is to pre-train \cite{hypergcl,tricl,kimhypeboy} independently on different domain-specific hypergraph structures to generate a hypergraph foundation model. 

However, this approach yields poor results, as shown in \cref{fig:neg-trans}. Pre-training on source domain hypergraph structures and then applying the model to target domain downstream tasks leads to negative transfer. This occurs because relational data from different domains exhibit significant structural pattern differences. Specifically, the lower part of \cref{fig:neg-trans} illustrates the underlying reason for performance degradation when transferring from Aminer-Text to other datasets. We plot the statistical distributions of hyperedge degrees and observe that Movielens-Text shares a distribution pattern similar to Aminer-Text, resulting in only a marginal performance drop of 0.64\%. In contrast, as the distributional discrepancy increases, the performance degradation becomes more pronounced. For instance, the hypergraph distribution of PPI-Text differs substantially from that of Aminer-Text, with distinct peak positions and fluctuation patterns, leading to a significant performance drop of 18.07\%. Consequently, how can multi-domain knowledge extraction and transfer be effectively achieved?

Existing methods typically concatenate structures directly, but we observe that this approach only performs adequately on small-scale datasets. It remains unclear whether large-scale knowledge extraction and transfer are feasible. Additionally, direct concatenation can result in the blending of domain-specific knowledge, as evidenced by the performance bottleneck observed in the experiments presented in \cref{exp:tab:clustering}. Thus, how can knowledge be transferred across domains while preserving the unique semantic information of each domain?

Moreover, research on hypergraph foundation models faces fundamental challenges at the data level. While there is an abundance of text-attributed graph data \cite{zhao2022learning,yan2023comprehensive,tape} —where each vertex is accompanied by textual descriptions—these descriptions facilitate better communication and collaboration between Graph Neural Networks (GNNs) and LM, spurring advancements in graph-based foundation models. In contrast, the text-attributed hypergraph (TAHG) dataset remains underexplored, representing a significant bottleneck for research at the intersection of hypergraphs and language models. The current lack of comprehensive TAHG datasets limits progress in this domain.

To address the aforementioned issues, we propose Hyper-FM: Hypergraph Foundation Model for Multi-Domain Hypergraph Knowledge Extraction. Specifically, for extracting vertex knowledge from hypergraph vertex features, we introduce Hierarchical High-Order Neighbor Guided Vertex Knowledge Embedding. This module hierarchically represents the hypergraph neighborhoods and defines a hierarchical domain prediction task to fine-tune the LM. This approach not only addresses the challenge of representing high-order domain information in large-scale hypergraphs but also injects domain information into the features extracted by the language model, achieving dual embedding of structural and semantic information. For extracting structural knowledge from hypergraphs, we present Hierarchical Multi-Hypergraph Guided Structural Knowledge Extraction. This method involves sampling sub-hypergraphs from different domain hypergraphs, clustering to construct a macro-hypergraph, and connecting different domain hypergraphs via bond vertices to form a Hierarchical Multi-Hypergraph. The sampling strategy effectively manages large-scale hypergraphs, while unsupervised clustering enhances collaborative learning among semantically similar vertices within the same domain. Additionally, the hierarchical structure with virtual vertices mitigates the direct impact of information from other domains, preventing the mixing of domain-specific information.

{
Furthermore, we have manually curated and constructed 11 TAHG datasets to facilitate cross-disciplinary research between HGNNs and LLMs. Experiments on these 11 hypergraph datasets demonstrate that our proposed foundation model outperforms baseline methods by an average of approximately 13.4\%, validating its effectiveness.} Additionally, we are the first to propose scaling laws for hypergraph computation. Our findings reveal that, unlike other domains, simply increasing the scale of data in terms of vertex and hyperedge counts does not enhance the foundational model's power, as illustrated in \cref{fig:num_sampling}. However, increasing the number of domains—thereby exposing the model to a wider variety of relational structures—significantly enhances the model's performance. This suggests a promising direction for future development. In summary, our contributions are threefold:
\begin{enumerate}
    \item {We curate and release 11 text-attributed hypergraph (TAHG) datasets, fostering the integration of hypergraph computation and language model research.}
    \item {We introduce the first hypergraph foundation model for extracting vertex and structural knowledge from multi-domain hypergraph data. Leveraging hierarchical representations and a sampling-based Hierarchical Multi-Hypergraph, our model achieves an average performance improvement of approximately 13.4\% across 11 TAHG datasets, demonstrating its effectiveness.}
    \item We propose the first scaling law for hypergraph foundation models, illustrating that increasing the number of vertices and hyperedges does not enhance model performance. Instead, augmenting the number of pre-training domains—thereby exposing the model to diverse relational structures—effectively improves the foundation model's power.
\end{enumerate}



\section{Related works}

\subsection{Foundation Models}

Foundation models have profoundly transformed NLP and CV. In NLP, BERT \cite{bert} revolutionized context-aware embeddings through bidirectional training, boosting tasks like question answering. GPT-2 \cite{gpt2} and GPT-3 \cite{gpt3} further demonstrated the power of generative pre-training, with GPT-3’s 175B parameters enabling impressive few-shot learning. T5 \cite{t5} unified NLP tasks under a text-to-text framework, while RoBERTa \cite{roberta} improved training efficiency by removing the next-sentence prediction objective and using larger datasets. In CV, models like DINO \cite{dido} leveraged self-supervised learning and knowledge distillation to learn rich visual representations without labels. SAM \cite{sam} advanced image segmentation, SegGPT \cite{seggpt} integrated segmentation and generation for complex tasks, and Visual ChatGPT \cite{visualgpt} combined GPT-3 with visual understanding for interactive multimodal applications. Collectively, these advances underscore the broad impact of foundation models in modern machine learning.

\subsection{Graph Foundation Models}
In recent years, several innovative graph foundation models have emerged, significantly advancing the field of graph data processing. InstructGLM \cite{instructglm} is a generative language model specifically designed for graph structures, enabling users to create and modify graphs through natural language instructions, thereby enhancing interactivity with graph data. GraphGPT \cite{graphgpt} combines graph neural networks with generative pre-trained transformers, focusing on generating and reasoning about graph data, which facilitates tasks such as graph description and generation through contextual information. LLaGA \cite{llaga} aligns language models with graph structures to improve representation learning by jointly training on both data types, boosting performance on tasks like vertex classification and link prediction. Finally, GALLM \cite{gallm} integrates graph information into language models, strengthening the connection between generated text and graph data, making it suitable for applications like knowledge graph construction and information retrieval.

\subsection{Hypergraph Neural Networks}

Hypergraph Neural Networks (HGNNs) have greatly advanced the modeling of high-order relationships in graph-structured data. HGNN \cite{hgnn} first introduced hypergraph convolution to capture higher-order vertex relations, while HGNN$^{+}$ \cite{hgnnp} enhanced neighbor aggregation via multilevel information fusion. HyperGCN \cite{hypergcn} implemented efficient hypergraph convolution layers, and HNHN \cite{hnhn} adopted hierarchical vertex representations for multilayer information propagation. HGAT \cite{hgat} integrated self-attention with hypergraphs to dynamically learn vertex and hyperedge importance. UniGNN \cite{unignn} provided a unified framework for diverse graph structures, improving model flexibility. HyperSAGE \cite{hypersage}, inspired by GraphSAGE \cite{graphsage}, introduced sampling-based aggregation for scalability. AllSet \cite{allset} leveraged set theory to optimize hypergraph representation, and 
DHHNN \cite{mei2025dhhnn} is a variational autoencoder-based framework for multimodal data integration and vertex classification, which combines hyperbolic geometry, dynamic hypergraphs and self-attention to address the static structure limitation of existing models.

\subsection{Pioneer in Hypergraph Foundation Models}

Recent advances have seen pioneering hypergraph foundation models emerge, expanding hypergraph learning across domains. HGFM \cite{han2025hypergraph} is a high-order correlation-driven model for brain disease diagnosis, learning an encoder via self-supervised pretraining on high-order structures and adapting to downstream tasks through few-shot fine-tuning. HYPER \cite{huang2025hyper} is a foundation model for inductive link prediction in knowledge hypergraphs, generalizing to novel entities and relations by encoding entities with their hyperedge positions. DMCL-HFM \cite{zhou2025dual} targets whole slide images, using vertex-hyperedge dual-masked modeling and contrastive learning for pretraining to integrate local and global dependencies. The hypergraph-driven landmark detection model \cite{dong2025hypergraph} supports cardiac function quantification, leveraging an adaptive dynamic hypergraph backbone and bidirectional hypergraph spatio-temporal decoder. Notably, these models are tailored to specific domain tasks (e.g., brain disease diagnosis), highlighting the lack of a universal hypergraph foundation model framework for multiple scenarios.

\section{Preliminaries}
\paragraph{Text-attributed Hypergraphs} A hypergraph can be represented as $ \gG = \{ \gV, \gE \} $, where $ \gV = \{ v_1, v_2, \ldots\} $ is the set of vertices and $ \gE = \{e_1, e_2, \ldots \}$ is the set of hyperedges. Typically, each vertex is associated with some textual descriptions, denoted as $ \gT = \{ T_1, T_2, \ldots \} $, where $ T_i $ is the textual description for vertex $ v_i $.

\paragraph{Hypergraph Neural Networks and Task Definition}
Hypergraph Neural Networks (HGNNs) are designed for feature representation learning on hypergraphs, facilitating feature convolution that is guided by the hypergraph structure. Classical HGNNs \cite{hgnn,hgnnp} consist of two stages of message passing: message propagation from vertices to hyperedges and from hyperedges back to vertices, which can be defined as:
\begin{equation}
    \mX^{(l+1)} = \sigma\left( \mD^{-\frac{1}{2}}_v \mH \mD^{-1}_e \mH^\top \mD^{-\frac{1}{2}}_v \mX^{(l)} \bm{\Theta}^{(l)} \right) ,
\end{equation}
where $\mD_v$ and $\mD_e$ are diagonal matrices representing the degree matrices of vertices and hyperedges, respectively. The matrix $\mH \in \{0, 1\}^{|\gV| \times |\gE|}$ is the incidence matrix of the hypergraph, where $\mH_{v,e} = 1$ indicates that vertex $v$ is part of hyperedge $e$. $N = |\gV|$ denotes the number of vertices. The vertex feature matrix is denoted by $\mX^{(l)} \in \sR^{N \times C^{(l)}}$, where $C^{(l)}$ represents the number of feature channels at layer $l$. The parameter $\bm{\Theta}^{(l)} \in \sR^{C^{(l)} \times C^{(l+1)}}$ consists of trainable weights for transforming features from layer $l$ to layer $l+1$. The activation function $\sigma$ introduces non-linearity into the model.

The objective of this study is focused on the vertex classification task. Each vertex $v$ is associated with a corresponding label $Y_v$, and the collection of all labels forms a label set $\gY$. The goal of the vertex classification task is to learn a mapping $\phi_\theta$ from vertex embeddings to their respective labels, thereby enabling the classification of vertices with unknown labels based on their learned representations.

\begin{figure*}[!ht]
    \centering
    \includegraphics[width=\linewidth]{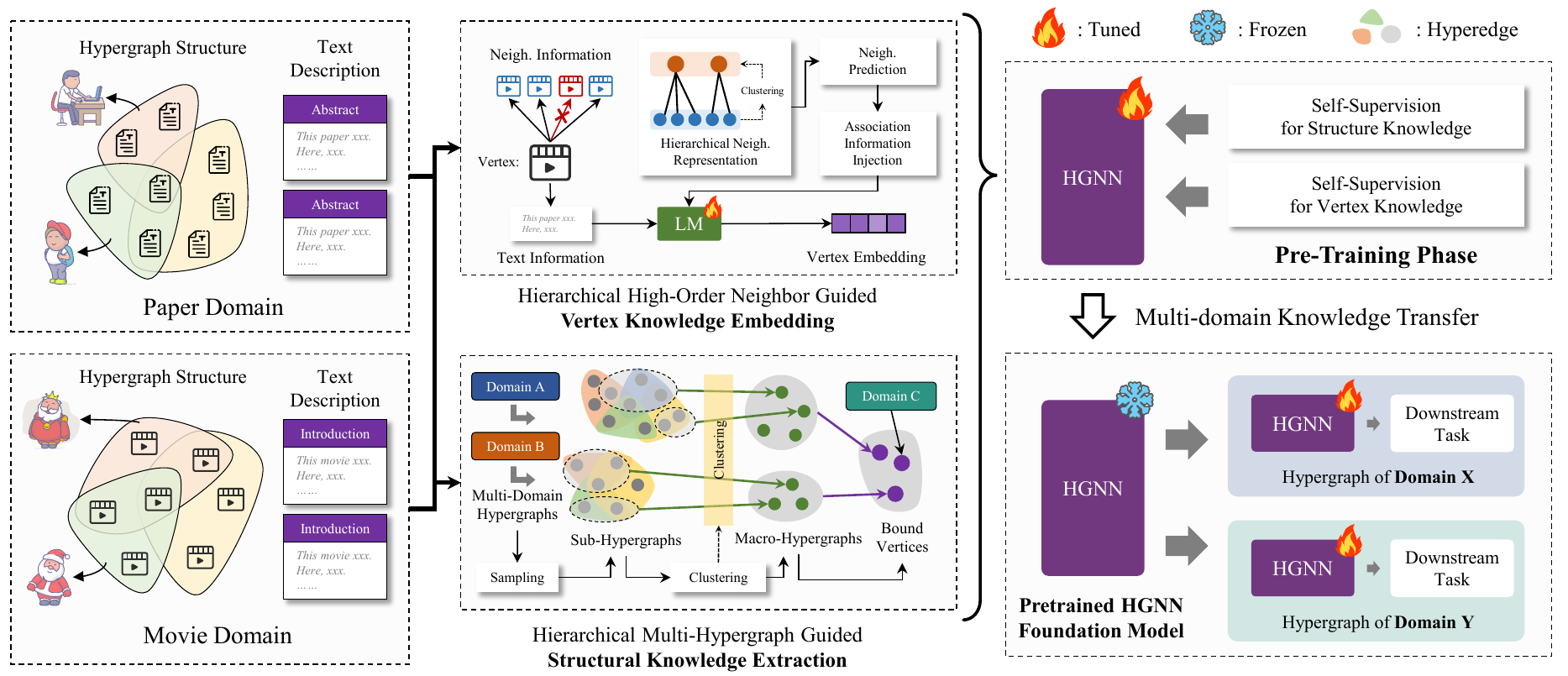}
    \caption{Pipeline of the proposed Hypergraph Foundation Model (Hyper-FM).}
    \label{fig:fw}
\end{figure*}

\paragraph{Hypergraph Foundation Models} 
In this study, we define Hypergraph Foundation Models to leverage multi-domain hypergraph data, specifically text-attributed hypergraphs, for obtaining pre-trained parameters of hypergraph neural networks through self-supervised learning. These pre-trained parameters serve as initialization for hypergraph neural networks applied to downstream tasks in other domains. By incorporating multi-domain hypergraph knowledge into the foundation model, we enhance the performance of hypergraph neural networks on target domain hypergraph data.
The training process of the Hypergraph Foundation Model consists of two phases: pre-training with multi-domain hypergraph data and fine-tuning for specific domain data. The pre-training phase is formally defined as:
\begin{equation}
    \theta^* = \argmin_{\theta} \sum_{(\gT_x, \gG_x) \in \gD^\text{src}} \gL_\text{pre} \left( f(\gT_x, \gG_x \mid \theta) \right),
\end{equation}
where $\theta^*$ denotes the optimized parameters of the Hypergraph Foundation Model after pre-training. The dataset $\gD^\text{src} = \{ (\gT_1, \gG_1), (\gT_2, \gG_2), \ldots \}$ comprises multi-domain hypergraph data, with each pair $(\gT_x, \gG_x)$ containing the textual attributes $\gT_x$ of the vertices and the structural information $\gG_x$ of the hypergraph. The textual attributes $\gT_x$ provide rich semantic information that complements the structural data captured by $\gG_x$. The loss function $\gL_\text{pre}$ corresponds to a predefined self-supervised hypergraph task designed to encode knowledge from multi-domain hypergraphs into the parameters. The function $f_\theta$ represents the hypergraph neural network model with trainable parameter $\theta$. Through self-supervised tasks, the model captures and integrates knowledge from diverse hypergraph domains into the foundational parameters $\theta^*$.

After pre-training, the foundation model is adapted to a specific target domain through fine-tuning. This process initializes the hypergraph neural network parameters with the pre-trained $\theta^*$ and optimizes them based on the target domain's data and labels. The fine-tuning process is defined as:

\begin{equation}
    \begin{aligned}
    \theta' = &\argmin_{\theta'} \sum_{(T_x, Y_x) \in (\gT^\text{tar}, \gY^\text{tar})} \gL_\text{down} \left( p_\theta(T_x, \gG^\text{tar}), Y_x \right) \\
    & \text{subject to} \quad \theta'^{(0)} \leftarrow \theta^*
    \end{aligned},
\end{equation}
where $(\gT^\text{tar}, \gG^\text{tar}, \gY^\text{tar})$ represents the target domain hypergraph data, including textual vertex attributes $\gT^\text{tar}$, hypergraph structure $\gG^\text{tar}$, and vertex labels $\gY^\text{tar}$. The textual attributes $\gT^\text{tar}$ provide semantic context for the vertices, while the structural information $\gG^\text{tar}$ captures the relationships between vertices via hyperedges. The label set $\gY^\text{tar} = \{ Y_1, Y_2, \ldots, Y_N \}$ consists of labels for each vertex in the target hypergraph, essential for supervised fine-tuning. The loss function $\gL_\text{down}$ pertains to the downstream vertex classification task, aiming to minimize the discrepancy between the predicted labels $p_\theta(T_x, \gG^\text{tar})$ and the ground true labels $Y^t$. The parameter $\theta'$ represents the fine-tuned parameters of the hypergraph neural network tailored to the target domain. The predictive function $p_\theta(T_x, \gG^\text{tar})$ maps the input textual attributes and hypergraph structures to label probabilities using the model parameters $\theta$. The initialization condition $\theta'^{(0)} \leftarrow \theta^*$ ensures that fine-tuning begins with the pre-trained foundation parameters, facilitating the transfer of multi-domain knowledge to the specific task.

By fine-tuning the foundation model on downstream tasks, we transform a generic hypergraph neural network into a specialized model optimized for the target domain. This approach effectively combines prior knowledge from multiple domains with task-specific data and supervision, overcoming the performance limitations inherent to single-domain models and achieving maximal performance improvements. The fine-tuned parameters $\theta'$ retain the generalizable knowledge acquired during pre-training while adapting to the specificities of the target domain, resulting in superior classification performance.
Through the proposed Hypergraph Foundation Model framework, we leverage multi-domain pre-training and domain-specific fine-tuning to develop robust and high-performing hypergraph neural networks across diverse application areas.

\section{Methodology}

In this section, we present the Multi-Domain Hypergraph Foundation Model (Hyper-FM). First, we overview its design and core objectives. Hyper-FM has two core modules: Hierarchical High-Order Neighbor Guided Vertex Knowledge Embedding (generates structure-aware vertex embeddings) and Hierarchical Multi-Hypergraph Guided Structural Knowledge Extraction (integrates cross-domain structural knowledge). We then describe its self-supervised multi-domain pre-training and domain-specific fine-tuning, and compare it with classic graph foundation models to show its advantages.

\subsection{Framework Overview}
\Cref{fig:fw} presents our Multi-Domain Hypergraph Foundation Model. Given multi-domain text-attributed hypergraphs, we first utilize the Vertex Knowledge Embedding module to extract structure-aware vertex features by constructing hierarchical neighbor labels, thereby integrating high-order structural information into the vertex representations. Next, the Structural Knowledge Extraction module constructs a hierarchical multi-domain hypergraph by adding bond vertices that connect hypergraphs from different domains, facilitating cross-domain information propagation. During pre-training, we randomly sample structures from the multi-domain hypergraphs and use the structure-aware embeddings as input features to train the hypergraph foundation model in a self-supervised manner. For downstream tasks, the pre-trained model parameters are initialized for the specific domain's hypergraph, allowing the model to leverage the acquired multi-domain knowledge and improve performance on the target domain hypergraph data.

\begin{figure*}[!ht]
    \centering
    \includegraphics[width=0.85\linewidth]{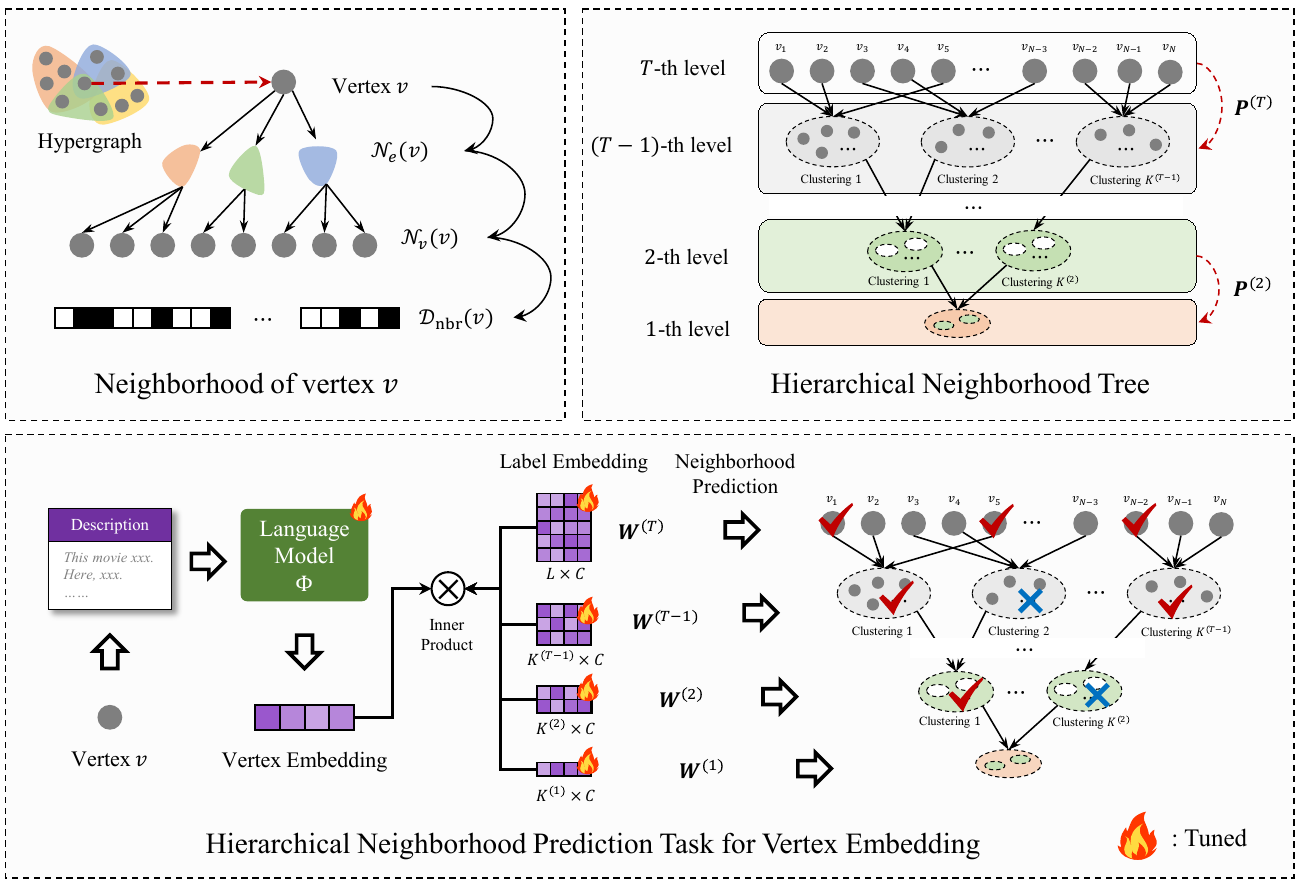}
    \caption{Illustration of the Hierarchical High-Order Neighbor Guided Vertex Knowledge Embedding Module.}
    \label{fig:tree}
\end{figure*}

\subsection{Hierarchical High-Order Neighbor Guided Vertex Knowledge Embedding}
\label{sec:v_ft}
In this subsection, we introduce the Vertex Knowledge Embedding module, as illustrated in \cref{fig:tree}. Initially, vertex textual descriptions are encoded using a Language Model (LM) to capture semantic information. However, this approach alone does not fully leverage the connectivity information inherent in hypergraph data. To address scenarios where a vertex's textual description is semantically ambiguous, we enhance its representation by incorporating descriptions from neighboring vertices. To facilitate this, we design a self-supervised neighborhood prediction task that guides the LM to extract features that consider vertex associations within the hypergraph. Direct pairwise link prediction is computationally inefficient, and predicting all neighbor-pairs would result in an excessively large and unmanageable label space. Therefore, we propose a hierarchical high-order neighborhood representation method that employs hierarchical label prediction. This approach enables the LM to discern and integrate high-order relational structures within the hypergraph, thereby enriching the vertex embeddings with comprehensive structural and semantic knowledge.

\subsubsection{High-Order Neighborhood Vectorization} 
We begin by defining the neighborhood vector within a hypergraph. For a given vertex $v$, we first identify its incident hyperedges $\gN_e(v)$. Subsequently, we collect all vertices $u$ connected by these hyperedges, forming a vertex set. This set is then vectorized into a binary vector $\gD_{\text{nbr}}(v) \in \{0, 1\}^{1 \times N}$, where $N$ represents the total number of vertices in the hypergraph. The vectorization process is formalized as:
\begin{equation}
    \gD_\text{nbr}(v) = \text{vec}\left(\{u \mid u \in e, \ e \in \gN_e(v), \ u \in \gV \}\right),
\end{equation}
where $\gD_{\text{nbr}}(v)_{[i]} = 1$ if vertex $v$ is connected to vertex $i$ via a hyperedge, and $0$ otherwise. Importantly, if two vertices $v_i$ and $v_j$ share identical neighborhoods, their vectors satisfy $\gD_{\text{nbr}}(v_i) = \gD_{\text{nbr}}(v_j)$.
Leveraging this vectorization, we define the training objective for the LM as follows:
\begin{equation}
     \big( \gD_\text{nbr}(v_i) \simeq \gD_\text{nbr}(v_j) \big) \propto \text{Similarity} \left( \phi(T_i), \phi(T_j) \right),
\end{equation}
where $\phi$ denotes the LM, and $T_i$ and $T_j$ are the textual descriptions of vertices $v_i$ and $v_j$, respectively. This objective ensures that vertices with similar neighborhood structures obtain comparable feature representations through the LM. Consequently, the LM is guided to incorporate high-order structural information from the hypergraph into the vertex embeddings, thereby enhancing both the semantic and structural integrity of the representations.


\subsubsection{Hierarchical Neighborhood Representation}
We transform neighborhood prediction into a multi-label classification problem. Specifically, for a vertex $v$, we aim to predict its connections to each vertex in the hypergraph using a binary vector of length $N$, where $N$ is the total number of vertices. Here, the number of classification labels is $L = |\mathcal{V}|$. However, when dealing with hypergraphs containing a large number of vertices, the resulting vectors become excessively long, rendering the model difficult to train efficiently. Inspired by Extreme Multi-label Classification  techniques\cite{zhang2021fast,prabhu2018parabel,khandagale2020bonsai}, we address this challenge by adopting a hierarchical representation of neighbor relationships through clustering, grouping similar neighborhoods into unified clusters, as shown in \cref{fig:tree}.

First, we define the fundamental feature representation for each label (\textit{i.e.}, the connection relationships of vertices) to facilitate clustering. For each label $x$, we aggregate the textual features of its associated vertices and normalize the resulting vector as follows:
\begin{equation}
\label{eq:gen_z}
    \mZ_x = \frac{\vv_x}{||\vv_x||}, \text{ where } \vv_x = \sum_{u \in \gN_v(x)} \psi(T_u), \forall x \in [L],
\end{equation}
where $\psi$ represents the basic text feature extraction network, such as TF-IDF\cite{tfidf}. It should be noted that $\psi$ only extracts initial textual features, which are not directly fed into downstream neural networks but rather serve as the basis for clustering to provide references for hierarchical domains. $T_u$ is the textual description of vertex $u$. The set $\gN_v(x)$ comprises vertices associated with label $x$. The matrix $\mZ \in \sR^{L \times C}$ serves as the initial representation matrix for the labels.

To manage large-scale hypergraphs where the number of vertices (and thus labels) $L$ can be both sparse and extensive, we employ a hierarchical clustering approach. At each clustering layer $t$, we project the label representations from the current level to the next coarser level using $k$-means clustering:
\begin{equation}
    \mP^{(t)} \leftarrow k\text{-means-clustering}(\mZ^{(t)}, K^{(t-1)})\revs{,}
\end{equation}
where $\mathbf{P}^{(t)} \in \{0, 1\}^{K^{(t)} \times K^{(t-1)}}$ is the projection matrix mapping clusters from layer $t$ to layer $t-1$, and $K^{(t)}$ denotes the number of clusters at layer $t$. The hierarchical process is repeated until the desired tree height $T$ is achieved, with the finest granularity at layer $T$ where $K^{(T)} = L$ and $\mZ^{(T)} = \mZ$. For coarser layers, the label representations are updated as follows:
\begin{equation}
    \mZ^{(t-1)} = \mZ^{(t)} \mathbf{P}^{(t)}\revs{,} 
\end{equation}
where $\mZ^{(t-1)} \in \sR^{K^{(t-1)} \times C}$ represents the aggregated label features at the $(t-1)$-th layer. Each row in $\mZ^{(t-1)}$ corresponds to a group of vertices from the previous layer, effectively capturing higher-order neighborhood structures.

Given that vertices with similar neighborhoods are likely to share cluster memberships, this hierarchical clustering approach not only reduces the dimensionality of the label space but also enhances training efficiency by grouping related labels. Furthermore, we define the neighborhood label matrix at each layer $t$ as:
\begin{equation}
    \mY^{(t)}_{v,x} = 
        \begin{cases}
            1 & \text{if vertex } v \text{ is connected to label } x, \\
            0 & \text{otherwise.}
        \end{cases}
\end{equation}

At the finest layer $T$, each label $x$ corresponds to a specific vertex. In higher layers, each label represents a cluster of vertices. To propagate label information to coarser layers, we update the neighborhood label matrix as follows:

\begin{equation}
    \mY^{(t-1)} = \text{bin}\left(\mY^{(t)} \mathbf{P}^{(t)}\right),
\end{equation}
where $\mY^{(t)} \in \{0, 1\}^{N \times K^{(t)}}$, and the $\text{bin}$ function binarizes the resulting matrix by setting all non-zero entries to 1. This hierarchical approach aggregates similar neighborhoods, creating a multi-level neighborhood representation. Consequently, structural information is progressively injected into the LM during training, enabling the handling of larger-scale hypergraphs and improving both encoding and computational efficiency.

\subsubsection{Association Information Injection}

As illustrated in \cref{fig:tree}, the hierarchical structuring of neighborhood within the hypergraph facilitates the effective utilization of neighborhood similarities and addresses the challenges associated with representing a large number of vertices. In this section, we describe the method for injecting association information from the hypergraph into the LM to generate structure-aware vertex embeddings. To fully exploit the cross-level associations in the hierarchical neighborhood representation, we adopt a hierarchical progressive learning approach.

Specifically, we begin at the first level and iteratively predict each vertex's associations with other vertices or vertex clusterings. At each level $t$, we first obtain the predicted associations for the $(t-1)$-th level, denoted as $\mY^{(t-1)}_\text{pred}$. To enhance training efficiency and robustness, inspired by beam search, we select the top $k$ most significant association labels for each vertex:
\begin{equation}
\label{eq:gen_y}
    \mY^{(t-1)}_\text{pred} = \text{Top}\left(\mW^{(t-1)\top} \Phi(\gT, \Theta), \ k\right) \revs{,}
\end{equation}
where $\mY^{(t-1)}_\text{pred} \in \{0, 1\}^{N \times K^{(t-1)}}$ indicates whether each vertex is associated with the top $k$ labels at level $t-1$. However, utilizing the entire vector for training would lead to an imbalance between positive and negative samples, resulting in suboptimal model performance. To mitigate this issue, we introduce a mask that selects a subset of samples for training at each level:

\begin{equation}
\label{eq:gen_s}
    \mS^{(t)} = \text{bin}(\mY^{(t-1)}_\text{pred}\mP^{(t)\top} + \mY^{(t-1)}\mP^{(t)\top})\revs{,}
\end{equation}
where $\mS^{(t)} \in \{0, 1\}^{N \times K^{(t)}}$ is the sample selection matrix. The first term, $\mY^{(t-1)}_\text{pred} \mP^{(t)\top}$, represents the associations predicted from the previous level, while the second term, $\mY^{(t)} \mP^{(t)\top}$, denotes the true associations at the current level. The $\text{bin}$ function converts the resulting matrix into a binary form, setting all non-zero entries to 1 and others to 0.

Based on this selection matrix, we define the optimization objective for learning the domain representations at level $t$ as follows:
\begin{equation}
\label{eq:gen_v_loss}
    \underset{\mW^{(t)}, \Theta}{\min} \sum\limits_{i=1}^{N} \sum\limits_{x:\mS^{(t)}_{i,x} \neq 0} \gL_\text{BCE}\left(\mY^{(t)}_{i,x}, \mW^{(t)\top}_{x} \Phi(\gT, \Theta)\right)\revs{,}
\end{equation}
where $\mW^{(t)}$ and $\Theta$ denote the embeddings of cluster labels at level $t$ and the parameters of the LM, respectively. We employ the Binary Cross-Entropy (BCE) loss function to compute the loss for each vertex-label association, facilitating the learning of connections between vertices and their corresponding domain clusters. This process is iteratively repeated from the first level to the final level $T$ until convergence, resulting in a LM that is aware of the hypergraph's structural associations. Consequently, the extracted vertex embeddings encapsulate both the semantic information from their textual descriptions and the structural knowledge from the hypergraph.
Finally, the structure-aware vertex embeddings $\mX \in \mathbb{R}^{N \times C}$ are generated as follows:
\begin{equation}
\label{eq:emb}
    \mX = \Phi(\gT, \Theta^*),
\end{equation}
where $\Theta^*$ represents the trained parameters of the LM, and $\gT$ includes the textual descriptions of the vertices within the hypergraph. These vertex embeddings are subsequently utilized for constructing hierarchical multi-domain hypergraphs and pre-training the multi-domain foundation model.

\subsection{Hierarchical Multi-Hypergraph Guided Structural Knowledge Extraction}
\begin{figure*}
    \centering
    \includegraphics[width=0.93\linewidth]{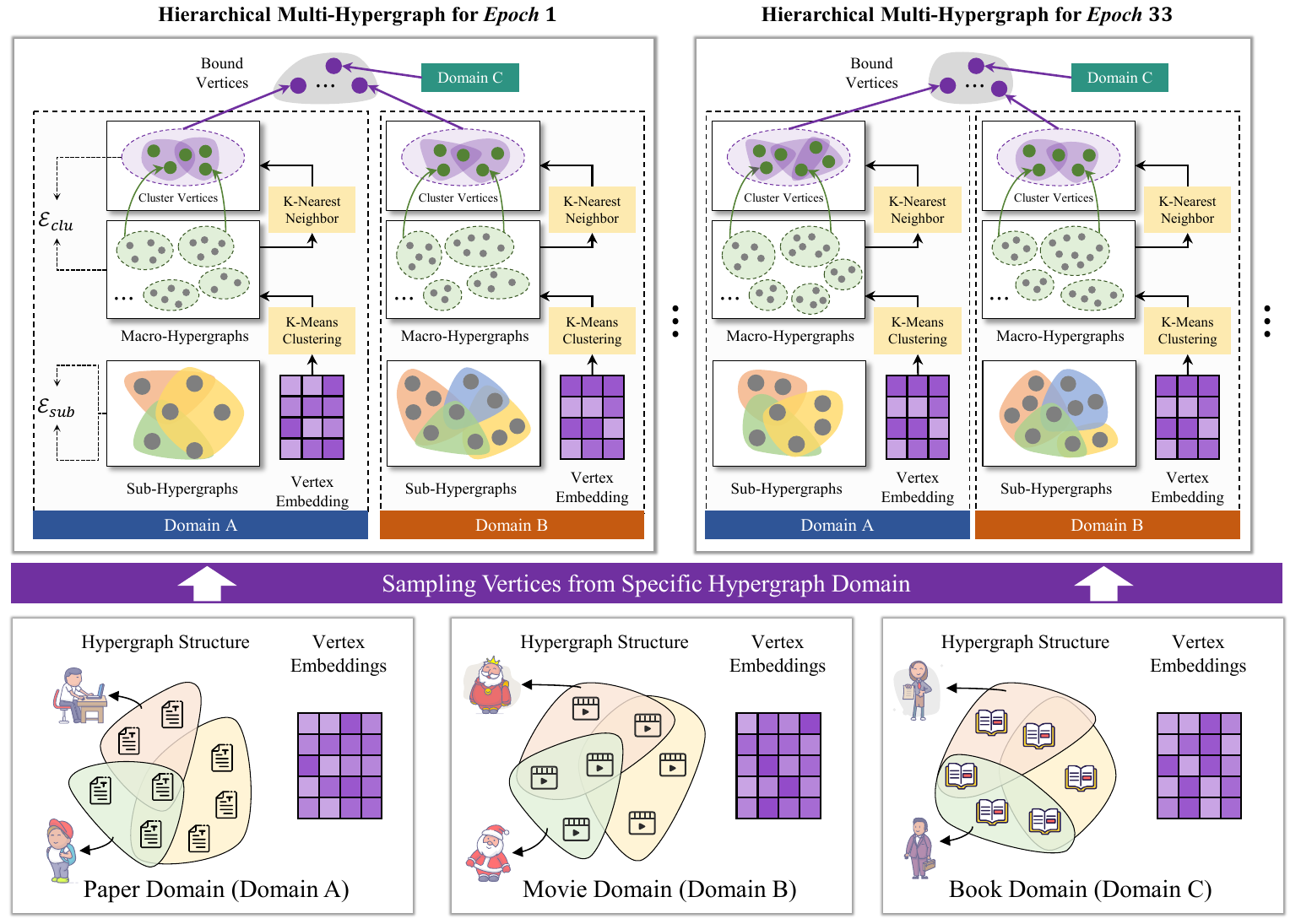}
    \caption{Illustration of sampling and building Hierarchical Multi-Hypergraphs for the pretaining} of hypergraph foundation model.
    \label{fig:cluster}
\end{figure*}

Different domains exhibit distinct structural association patterns within their respective hypergraphs. Preliminary experiments (as shown in \cref{fig:neg-trans}) indicate that isolated pretraining for each domain struggles to bridge the distributional discrepancies inherent in multi-domain hypergraph data. To overcome this limitation, we construct a hierarchical multi-domain hypergraph that integrates structural knowledge across domains, as shown in \cref{fig:cluster}, thereby dismantling the barriers imposed by domain-specific structural constraints.

\subsubsection{Multi-Domain Hypergraphs Sampling}
Given hypergraph data from $m$ distinct domains, we employ an independent sampling strategy for each domain-specific hypergraph to extract representative substructures. Specifically, for each domain $a \in \{1, 2, \ldots, m\}$ with its corresponding hypergraph $\gG^a = \{\gV^a, \gE^a\}$, we perform the following sampling process to obtain the sub-hypergraphs $\gG^a_\text{sub}$:
\begin{equation}
    \{\gG^1_\text{sub}, \gG^2_\text{sub}, \ldots, \gG^m_\text{sub}\} \xleftarrow{\text{sampling}} \{\gG^1, \gG^2, \ldots, \gG^m\}.
\end{equation}

For a specific domain $a$, the sub-hypergraph $\gG^a_\text{sub} = \{\gV^a_\text{sub}, \gE^a_\text{sub}\}$ is constructed as follows:
\begin{equation}
\begin{aligned}
    \gG^a_\text{sub} &= \{\gV^a_\text{sub}, \gE^a_\text{sub}\} \quad \text{where} \\
    &\gV^a_\text{sub} = \{\text{BFS}(v) \mid v = \text{random\_select}(\gV^a)\}, \\
    &\gE^a_\text{sub} = \{\text{filter}(e, \gV^a_\text{sub}) \mid e \in \gE^a\}.
\end{aligned}
\end{equation}
where $\gV^a_\text{sub}$ and $\gE^a_\text{sub}$ denote the sampled vertex and hyperedge sets, respectively. The sampling process begins by randomly selecting a vertex $v$ from the original vertex set $\gV^a$ of domain $d$. Starting from this vertex, a breadth-first search (BFS) is conducted to expand the vertex set, adding neighboring vertices iteratively until a specified number of vertices is reached. This ensures that the sampled vertex set $\gV^a_\text{sub}$ captures the local structural characteristics of the original hypergraph.
Subsequently, the hyperedge set $\gE^a_\text{sub}$ is derived by filtering the original hyperedges $\gE^a$ to retain only those hyperedges where all constituent vertices are present in $\gV^a_\text{sub}$. Formally, for each hyperedge $e \in \gE^a$, it is included in $\gE^a_\text{sub}$ if and only if $e \subseteq \gV^a_\text{sub}$. This filtering step ensures that the sub-hypergraph $\gG^a_\text{sub}$ maintains meaningful high-order relationships inherent in the original hypergraph while reducing complexity.
This multi-domain sampling approach effectively captures the nuanced structural patterns of each domain's hypergraph, facilitating the subsequent clustering and integration steps. 

\subsubsection{Multi-Domain Hypergraphs Clustering}

For each sampled sub-hypergraph structure $\gG^a_\text{sub}$, we  extract the feature matrix $\mX_\text{sub}^a$ corresponding to the subset of vertices based on embeddings $\mX^a$ from \cref{eq:emb}. We then apply $k$-means clustering to the embeddings to partition the vertices into $k$ distinct clusters as follows:
\begin{equation}
    \{c_1^a, c_2^a, \cdots, c_k^a \} \leftarrow \revs{k\text{-means-clustering}}(\mX_\text{sub}^a, k),
\end{equation}
where $k$ is a hyperparameter that determines the number of clusters, and each $c_j^a$ represents a cluster containing a subset of vertices from domain $a$. It is important to note that each cluster $c_j^a$ is a collection of vertices.

Subsequently, for each cluster $c_j^a$, we construct a virtual vertex to represent the cluster. The feature vector of this virtual cluster vertex, denoted as $\vx^\text{clu}_{a, c_j}$, is obtained by aggregating the feature vectors of all vertices within the cluster:
\begin{equation}
    \vx^\text{clu}_{a, c_j} = \frac{1}{|c^a_j|}\sum_{u \in c^a_j} \vx_u.
\end{equation}
This aggregation ensures that the virtual vertex encapsulates the collective characteristics of the vertices within its cluster, providing a meaningful representation for higher-level structural analysis.

To capture high-order associations between these cluster vertices, we employ a $k$-nearest neighbors (k-NN) approach based on the aggregated cluster features $\mX^a_\text{clu}$. This process constructs connections between clusters that exhibit similar feature representations, thereby encoding the structural relationships at a higher abstraction level. The hyperedges are then formed by combining these k-NN connections with the original hyperedges derived from the clustering process:
\begin{equation}
\label{eq:gen_clu_hg}
\left\{
\begin{aligned}
    \gV^a_\text{clu} &= \{ v^\text{clu}_{a,1},  v^\text{clu}_{a,2}, \cdots,  v^\text{clu}_{a,k} \}, \\
    \gE^a_\text{clu} &= \text{knn}(\mX^a_\text{clu}, k_c) \cup \{c^a_1, \cdots, c^a_k \}
\end{aligned}
\right. ,
\end{equation}
where $\gV^a_\text{clu}$ denotes the set of virtual cluster vertices for domain $a$, and $\gE^a_\text{clu}$ represents the set of high-level hyperedges. The function $\text{knn}(\mX^a_\text{clu}, k_c)$ generates hyperedges based on the $k_c$ nearest neighbors in the cluster feature space, while $\{c^a_1, \cdots, c^a_k \}$ includes the original clustered hyperedges.
This clustering strategy effectively reduces the complexity of the hypergraph by grouping similar vertices, enabling the extraction of higher-order structural patterns within each domain. By constructing virtual vertices and establishing high-order associations, we enhance the representation of multi-domain hypergraphs, facilitating more efficient training and improved structural knowledge extraction in subsequent model stages.

\subsubsection{Hierarchical Multi-Hypergraph Construction}

In the subsequent step, we integrate the sampled sub-hypergraphs and their corresponding clusters from multiple domains to form the final hierarchical multi-domain hypergraph. Specifically, for each domain, a bond vertex is constructed and connected to its respective cluster vertices, thereby generating domain-specific hyperedges. Additionally, bond vertices across different domains are fully interconnected to facilitate the transfer of structural information between domains. The construction process is formalized as follows:
\begin{equation}
\left\{
\begin{aligned}
    \gV_\text{bond} &= \{ v^\text{bond}_a, v^\text{bond}_b, \cdots,  v^\text{bond}_m\} \\
    \gE_\text{bond} &= \{ (v^\text{bond}_a, v^\text{bond}_b,  \cdots,  v^\text{bond}_m), \\ 
    &\qquad (v^\text{bond}_a, v^{\text{clu}}_{a,1},  v^\text{clu}_{a,2}, \cdots,  v^\text{clu}_{a,k}), \\
    &\qquad (v^\text{bond}_b, v^{\text{clu}}_{b,1},  v^\text{clu}_{b,2}, \cdots,  v^\text{clu}_{b,k}), \\
    &\qquad \qquad \cdots \\
    &\qquad (v^\text{bond}_m, v^{\text{clu}}_{m,1},  v^\text{clu}_{m,2}, \cdots,  v^\text{clu}_{m,k}) \}
\end{aligned}
\right.
\end{equation}
where $\gV_\text{bond}$ denotes the set of bond vertices, with each $v^\text{bond}_d$ corresponding to domain $d \in \{a, b, \ldots, m\}$. The set $\gE_\text{bond}$ comprises two types of hyperedges: 
\begin{itemize}
    \item A hyperedge connecting all bond vertices across different domains, facilitating cross-domain information flow.
    \item Hyperedges that connect each bond vertex to its respective domain's cluster vertices, thereby consolidating the high-level structural information within each domain.
\end{itemize}

Next, for each bond vertex within a domain, we aggregate the features of its associated cluster vertices to form the bond vertex’s feature vector:
\begin{equation}
    \vx^\text{bond}_d = \frac{1}{k} \sum_{j=1}^{k} \vx^\text{clu}_{d, c_j} \revs{,}
\end{equation}
where $\vx^\text{bond}_d$ represents the feature vector for bond vertex $d$, and $\vx^\text{clu}_{d, c_j}$ denotes the feature vector of the $j$-th cluster vertex within domain $d$. This aggregation process ensures that each bond vertex encapsulates the aggregated structural features of its domain's clusters.
Subsequently, we compile the features of all bond vertices across domains into a bond vertex feature matrix $\mX_\text{bond} \in \mathbb{R}^{m \times C}$ by stacking the individual bond vertex features as follows:
\begin{equation}
    \mX_\text{bond} = \vx^\text{bond}_a || \vx^\text{bond}_b || \cdots || \vx^\text{bond}_m \revs{.}
\end{equation}

Finally, we integrate all components to construct the hierarchical multi-domain hypergraph $\gG_\text{M}$ as defined below:
\begin{equation}
\label{eq:gen_hmhg}
    \left\{
    \begin{aligned}
        \gG_\text{M} &= \{\gV_\text{M}, \gE_\text{M} \}, \\
        \gV_\text{M} &= \gV_\text{bond} \cup \bigcup_{d=1}^{m} \left( \gV_\text{sub}^d \cup \gV_\text{clu}^d \right), \\
        \gE_\text{M} &= \gE_\text{bond} \cup \bigcup_{d=1}^{m} \left( \gE_\text{sub}^d \cup \gE_\text{clu}^d \right), \\
        \mX_\text{M} &= \mX_\text{bond} || \mX_\text{sub}^a || \mX_\text{clu}^a || \cdots || \mX_\text{sub}^m || \mX_\text{clu}^m
    \end{aligned}
    \right.
\end{equation}
where $\gV_\text{M}$ represents the unified set of vertices, including bond vertices, sampled sub-hypergraph vertices, and cluster vertices across all domains. $\gE_\text{M}$ comprises all hyperedges, integrating bond hyperedges with sub-hypergraph and cluster hyperedges from each domain. $\mX_\text{M}$ denotes the comprehensive feature matrix, concatenating the bond vertex features with those of sampled sub-hypergraph vertices and cluster vertices.

This hierarchical multi-domain hypergraph structure is derived through the sampling of multiple domain-specific hypergraphs, thereby supporting the training of a multi-domain foundation model. During training, sampling can be performed in each epoch to ensure comprehensive coverage of the hypergraph information. Notably, the virtual vertices within the hierarchical multi-domain hypergraph, comprising both cluster vertices and bond vertices, are characterized by aggregated features rather than trainable embeddings. This design choice enhances the transferability of the model, enabling the construction of arbitrary multi-domain hypergraphs and supporting effective learning across diverse domains.

\subsection{Pretraining on Multi-Domain Hypergraphs}
Here, we describe the pretraining procedure for obtaining the parameters of the hypergraph foundation model using multi-domain hypergraph data, as outlined in \Cref{alg:pre-train}. 

\begin{algorithm}
\caption{Toward Multi-Domain Hypergraph Pre-training.}\label{alg:pre-train}
\begin{algorithmic}[1]
\Require Multi-domain text-attributed hypergraph datasets $\{(\gT_1, \gG_1), (\gT_2, \gG_2), \cdots, (\gT_m, \gG_m)\}$, text description of vertices $\gT = \{T_1, T_2, \cdots, T_N\}$, hypergraph neural networks with trainable parameters $\theta$, LM $\{\Theta_j\}^j$ initialize raw neighborhood (label) embedding $\mZ$ by \cref{eq:gen_z}, initialize neighbor embedding $\mW$, neighborhood number at each level of hierarchical neighborhood representation $\gK = \{K^{(1)}, K^{(2)}, \cdots, K^{(t)}\}$, learning rate $\alpha$.
\Ensure Pre-trained parameters $\theta^*$ of foundation model.
\State \textit{// Extract vertex features} 
\For{$j$ in $1, 2, 3, \cdots, m$}
\State \textit{// Build hierarchical neighborhood representation}
\State $\{\mP^{(t)}\}_{t=1}^{T}$ $\leftarrow$ \revs{$k$-means-clustering$(\mZ, \gK)$}
\State $\{\mY^{(t)}\}_{t=1}^{T}$ $\leftarrow$ $\text{binary}(\mY^{(t)}\mP^{(t)})$
\State \textit{// Training for each neighborhood level}
\For{$t$ in $1, 2, 3, \cdots T$}
\State $\mY^{(t-1)}_\text{pred} \leftarrow \text{Top}(\mW^{(t-1)\top}\Phi_j(\gT_j, \Theta_j), k_t)$
\State $\mS^{(t)} = \text{binarize}(\mY^{(t-1)}_\text{pred}\mP^{(t)\top} + \mY^{(t-1)}\mP^{(t)\top})$
\State update $\mW^{t}, \Theta_j$ by \cref{eq:gen_v_loss}
\EndFor
\State \textit{// Generate vertex features for domain $j$}
\State $\mX_j \leftarrow \Phi_j(\gT_j, \Theta_j)$
\EndFor
\State Get multi-domain vertex feature set: $\{\mX_1, \cdots, \mX_m\}$
\State \textit{// Pretraining}
\While{not converged}
\State \textit{// Sample multi-domain hypergraphs}
\State $\{(\gG^j_\text{sub}, \mX^j_\text{sub})\}_{j=1}^{m}$ $\leftarrow$ sampling$(\{(\gG^j, \mX_j)\}_{j=1}^{m})$
\State \textit{// Cluster vertices for each domain}
\For{$j$ in $1, 2, 3, \cdots, m$}
\State $\{c_1^j, c_2^j, \cdots, c_k^j \}$ \revs{$\leftarrow$ k-means-clustering$(\mX^j_\text{sub}, k)$}
\State $(\gV^j_\text{clu}, \gE^j_\text{clu})$ $\leftarrow$ by \cref{eq:gen_clu_hg}
\EndFor
\State \textit{// Build hierarchical multi-domain hypergraph}
\State $(\gG_M, \mX_M)$ $\leftarrow$ by \cref{eq:gen_hmhg}
\State \textit{// Training HGNN for one epoch}
\State $\theta \leftarrow \theta - \alpha \nabla_\theta \gL_\text{pre}(\gG_M, \mX_M)$
\EndWhile
\State $\theta^* \leftarrow \theta$
\State \Return $\theta^*$
\end{algorithmic}
\end{algorithm}

Given a collection of $m$ multi-domain text-attributed hypergraph datasets $\{(\gT_1, \gG_1), (\gT_2, \gG_2), \cdots, (\gT_m, \gG_m)\}$, we begin by extracting structure-aware vertex features for each domain-specific hypergraph. This extraction leverages the textual descriptions associated with each vertex, denoted by $\gT = \{T_1, T_2, \cdots, T_N\}$, which are processed through domain-specific LM $\{\Theta_j\}_{j=1}^m$. 
During the pretraining process, each epoch begins with randomly sampling sub-hypergraphs from each domain to obtain $\{(\gG^j_\text{sub}, \mX^j_\text{sub})\}_{j=1}^{m}$. For each domain $j$, the sampled vertex features $\mX^j_\text{sub}$ undergo $k$-means clustering to partition the vertices into $k$ distinct clusters, resulting in clusters $\{c_1^j, c_2^j, \cdots, c_k^j\}$. Each cluster $c_c^j$ comprises a set of vertices, and a corresponding virtual cluster vertex is created with its feature vector $\vx^\text{clu}_{j, c}$ derived by averaging the features of all member vertices within the cluster. Subsequently, the hierarchical multi-domain hypergraph $\gG_\text{M}$ is constructed by integrating the sampled sub-hypergraphs and their respective clusters using the methodology described in \cref{sec:v_ft}. The vertex feature matrix $\mX_\text{M}$ is formed by concatenating the features of bond vertices, sampled sub-hypergraph vertices, and cluster vertices across all domains.

The constructed hierarchical multi-domain hypergraph $(\gG_\text{M}, \mX_\text{M})$ is then input into the HGNN for training. The model parameters $\theta$ are updated by minimizing the pretraining loss $\gL_\text{pre}$, which comprises both structural and feature-based self-supervised components as defined in Equation \cref{eq:ssl}:
\begin{equation}
\label{eq:ssl}
    \gL_\text{pre} = \gL_\text{stru}(\gG_\text{M}, \tilde{\gG}_\text{M} \mid \theta) + \gL_\text{feat}(\mX_\text{M}, \tilde{\mX}_\text{M} \mid \theta),
\end{equation}
where, the structural self-supervised loss $\gL_\text{stru}$ is implemented using contrastive learning techniques applied to augmented structural views of the hypergraph, such as those employed in HyperGCL \cite{hypergcl}.  Following the design in SS-HT\cite{feng2025self}, we adopt both structural and feature-level self-supervised losses. Since the magnitudes of $\gL_\text{stru}$ and $\gL_\text{feat}$ are comparable, we simply aggregate them without additional weighting. This encourages the model to learn robust structural representations by distinguishing between different structural augmentations. The feature self-supervised loss $\gL_\text{feat}$ involves masking portions of the vertex features and reconstructing them based on domain-specific information, similar to the approach used in GraphMAE \cite{hou2022graphmae}. This mechanism ensures that the model effectively captures both the structural dependencies and the semantic features of the vertices.

The pretraining process iterates until convergence, resulting in the optimized parameters $\theta^*$ of the hypergraph foundation model. These pre-trained parameters encapsulate comprehensive structural and feature information across multiple domains, thereby enhancing the model's capability to generalize and perform robustly on downstream hypergraph tasks.

\subsection{Applying Knowledge to Downstream Hypergraphs}
In this section, we elucidate the methodology for leveraging the pre-trained hypergraph foundation model on hypergraph data from new domains. Building upon the previously established pretraining steps, which utilized multi-domain hypergraphs to learn the foundational parameters $\theta$ of the HGNN, we demonstrate how these parameters can be effectively transferred to target domain hypergraphs to enhance their performance on specific tasks.

To apply the foundation model to a target domain, we begin by initializing the HGNN for the target hypergraph with the pre-trained parameters $\theta$. This initialization imbues the target model with the structural and feature-based insights acquired from the diverse multi-domain hypergraphs during pretraining. Subsequently, the model architecture is adapted to the specific downstream task by appending an appropriate prediction head tailored to the task's requirements.
For vertex classification tasks, an additional MLP layer is attached to each vertex's embedding. This MLP layer is responsible for mapping the enriched vertex embeddings to the corresponding label predictions, thereby facilitating accurate classification based on the learned representations. In the case of hypergraph classification tasks, a global pooling layer is employed to aggregate the vertex embeddings into a single hypergraph-level embedding. This aggregated representation is then passed through a classification layer to determine the hypergraph's class label, effectively capturing the overarching structural and semantic attributes of the hypergraph.
For link prediction tasks, the approach involves directly utilizing the vertex embeddings to compute the probability of the existence of hyperedges. This is achieved by aggregating the embeddings of vertices that constitute a potential hyperedge and subsequently applying a suitable function to estimate the likelihood of their co-occurrence within a hyperedge. This strategy leverages the pre-trained embeddings to discern intricate relationships between vertices, thereby enhancing the model's predictive capabilities.

Throughout the fine-tuning process, the pre-trained parameters $\theta$ serve as a robust foundation, enabling the model to generalize effectively across various downstream tasks by incorporating prior knowledge from multiple domains. Empirical evaluations will subsequently demonstrate that the hypergraph foundation model not only encapsulates valuable prior information from multi-domain hypergraphs but also significantly elevates the performance benchmarks within target hypergraph domains for the vertex classification task. This transferability underscores the efficacy of the pretraining strategy in fostering versatile and high-performing hypergraph neural networks applicable to a wide range of real-world applications.

\subsection{Discussions}

We compare the proposed Hyper-FM with classical graph foundation models, focusing on core design differences and framework positioning, starting with Graph COordinators for PrEtraining (GCOPE)\cite{zhao2024all}.

First, in vertex feature generation: GCOPE uses BoW \cite{BoW} or TF-IDF \cite{tfidf} for feature extraction, then unifies dimensions via SVD \cite{svd}, failing to fully leverage advanced language models (LMs) and ignoring inherent data relationships. Hyper-FM, by contrast, adopts LMs to extract fine-grained textual features, and fine-tunes LMs via hierarchical domain representations and domain prediction tasks, integrating structural information into features.
Second, in multi-domain association structure generation: GCOPE concatenates multi-domain structures using virtual vertices connected to all domain vertices, causing information confusion among semantically distinct vertices. Hyper-FM constructs hierarchical multi-domain hypergraphs via clustering, strengthening connections between semantically similar vertices and linking domain vertices to clustered ones, enabling orderly cross-domain knowledge transfer.
Third, in scalability and transferability: GCOPE’s direct structure concatenation struggles with large-scale hypergraphs, and its domain-specific trainable virtual vertex embeddings require full retraining for new domains. Hyper-FM samples substructures per epoch to handle large-scale data, and generates virtual cluster/bond vertex features via aggregation, avoiding extra trainable parameters and supporting seamless new domain integration.

Beyond GCOPE, we further compare Hyper-FM with LLaGA\cite{llaga} and GraphGPT\cite{graphgpt}, which represent an LLM-centered paradigm. LLaGA reorganizes graph nodes into structure-aware sequences and maps them to LLM token space via a projector, adapting graph data to LLM inputs with LLM as the core. GraphGPT integrates LLMs with graph knowledge through graph instruction tuning, including text-graph grounding and dual-stage tuning, enabling LLMs to understand graph structures. Both models rely on LLMs to capture graph information by ``tokenizing'' graphs. Hyper-FM, however, takes GNNs as the core framework; LMs only serve to align textual features across domains, assisting GNNs in learning better hypergraph knowledge—representing an essentially different design paradigm.

Overall, Hyper-FM outperforms GCOPE in vertex feature generation, multi-domain structure construction, and scalability. It also differs fundamentally from LLaGA/GraphGPT in framework positioning, setting a new benchmark for hypergraph foundation models.
\section{Experiments}
In this section, we collect and build eleven text-attributed hypergraph datasets and conduct experiments on them to demonstrate the effectiveness of Hyper-FM. 

\subsection{Text-Attributed Hypergraph Datasets}
This section introduces the text-attributed hypergraph datasets (TAHG) developed to support the Hypergraph Foundation Model. Recognizing that most existing hypergraph datasets feature vertex attributes primarily comprised of pre-extracted visual features or Bag-of-Words vector encodings, which are incompatible with large-scale models, we have manually constructed eleven text-attributed hypergraph datasets. These datasets are categorized into four distinct classes to encompass a wide range of application domains.

The \textit{Citation Network Datasets} encompass Cora-CA-Text, Cora-CC-Text, Pubmed-CA-Text, Pubmed-CC-Text, Aminer-Text, Arxiv-Text, and OGBN-Arxiv. In these datasets, vertices represent papers with titles and abstracts serving as textual descriptions. For Cora and Pubmed, we collect textual descriptions based on the original paper IDs since this information is not available in current databases; Hyperedges are generated using co-author and co-citation relationships to capture the intricate interconnections within academic communities. For Arxiv-Text, we scrape data from the Arxiv repository, aggregating computer science papers published from 2010 to 2023, selecting data vertices from five major categories (Computer Vision, Computation and Language, Robotics, Computers and Society, Cryptography and Security, Software Engineering), and forming hyperedges via co-citation relationships. Additionally, for Aminer-Text, we scrape data from the Aminer website and generate hyperedges based on co-author relationships. OGBN-Arxiv is an additional large-scale dataset directly adopted from public resources\footnote{\url{https://ogb.stanford.edu/docs/nodeprop/}}. It is a directed citation network of Computer Science arXiv papers indexed by MAG \cite{wang2020microsoft}, where each vertex denotes an arXiv paper with textual descriptions consisting of the original paper's title and abstract. Each directed edge represents a citation relationship, and its task is 40-class subject area prediction (e.g., cs.AI, cs.LG).

The \textit{Visual Media Datasets} include the Movielens-Text dataset and IMDB-Text, where vertices correspond to movies enriched with textual descriptions, including the movie titles and introduction. For IMDB-Text, we scrape all movie data from the IMDB website for the years 2010 to 2023 and select data from five major film categories: Comedy, Horror, Documentary, Sci-Fi, and Animation. Hyperedges are generated based on co-director relationships to enable a detailed examination of film collaborations. For Movielens-Text, we utilize the original data from the ml-latest dataset and construct hyperedges based on co-writer relationships among films, facilitating the exploration of narrative connections across different movies.

As for the \textit{Book Domain}, we build the GoodBook-Text dataset, which models bibliographic information where vertices represent books accompanied by textual metadata. We utilize the original data from the GoodBook dataset and collect book names and details as textual descriptions for each vertex, allowing for a clear representation of books within the network. 

As for the \textit{Protein Domain}, we build the PPI-Text dataset, where vertices represent proteins and hyperedges denote protein-protein interactions. These hyperedges are supplemented with textual annotations that provide functional and structural insights, and the textual descriptions for each vertex consist of the protein sequences. This comprehensive collection of datasets facilitates the study of complex relationships across various domains, enhancing the applicability of our Hypergraph Foundation Model.

The statistics of these datasets are given in \cref{exp:tab:dataset}. By incorporating rich textual attributes, these datasets enhance the capability of Hyper-FM to leverage semantic information across multiple domains, thereby facilitating more effective knowledge extraction and transfer learning in downstream tasks.

\begin{table}[!ht]
\centering
\begin{threeparttable}
\caption{Statistics results of text-attributed hypergraph datasets.}
\label{exp:tab:dataset}
\begin{tabularx}{\linewidth}{lccRRR}
\toprule 
Dataset & \#Label & \#Text & \#Avg. & \#V & \#E \\ 
\midrule
Aminer-Text & 3 & Title+Abstract & 215.45 & 8,226 & 4,050 \\
Cora-CA-Text & 7 & Title+Abstract & 155.89 & 2,708 & 1,922 \\
Cora-CC-Text & 7 & Title+Abstract & 155.89 & 2,708 & 2,161 \\
Pubmed-CA-Text & 3 & Title+Abstract & 249.16 & 19,717 & 28,154 \\
Pubmed-CC-Text & 3 & Title+Abstract & 249.16 & 19,717 & 10,547 \\
Arxiv-Text & 5 & Title+Abstract & 194.10 & 22,886 & 15,478 \\
Movielens-Text & 5 & Name+Intro. & 43.26 & 18,479 & 8,271 \\
IMDB-Text & 5 & Name+Intro. & 57.43 & 34,619 & 13,290 \\
GoodBook-Text & 4 & Name+Detail & 182.53 & 5,834 & 7,920 \\
PPI-Text & 3 & Sequence & 587.84 & 401 & 381 \\
OGBN-Arxiv & 40 & Title+Abstract & 170.36 & 169,343 & 168,537 \\
\bottomrule
\end{tabularx}
\begin{tablenotes}
\footnotesize
\item \#V denotes ``Number of vertices'', \#E denotes ``Number of hyperedges'', \#Classes denotes ``Number of classes'', \#Text denotes ``Type of text description'', \#Avg. denotes ``Average length of vertex text descriptions''. ``Intro.'' is short for ``Introduction''.
\end{tablenotes}
\end{threeparttable}
\end{table}

\subsection{Experimental Settings}
\paragraph{Compared Methods} 
In our experiments, we evaluate the performance of the proposed \textit{Hyper-FM+Finetune} against a range of baseline methods to demonstrate its effectiveness. Firstly, we compare our approach with supervised methods that utilize the same labeled data. This includes a Multi-Layer Perceptron (MLP) with an identical number of layers to our foundation model, serving as a straightforward baseline that processes vertex features without leveraging hypergraph structures. Additionally, we benchmark against classical hypergraph neural networks such as HGNN \cite{hgnn}, HGNN$^+$ \cite{hgnnp}, HNHN \cite{hnhn}, and UniGCN \cite{unignn}, which are well-established models in hypergraph representation learning. To assess the efficacy of our proposed multi-domain joint pretraining strategy, we also compare it with pretraining-based methods. Specifically, we employ Isolated Pretraining with Finetuning (IP+Finetune), which utilizes the same domain-specific hypergraph data for pretraining as our approach. However, IP+Finetune adopts a naive sequential pretraining strategy, where the model is pre-trained on one domain's hypergraph data before proceeding to the next, without exploiting potential synergies between different domains during the pretraining phase.

Our proposed method, Hyper-FM+Finetune, embodies the Hypergraph Foundation Model strategy. For hypergraph pretraining, we utilize two self-supervised techniques: HyperGCL \cite{hypergcl}, a representative contrastive learning framework tailored for hypergraphs, and SS-HT\cite{feng2025self}, which leverages the reconstruction of hypergraph vertex features as a self-supervised objective. The backbone of our foundation model incorporates both HGNN and HGNN$^+$ architectures. In the HyperGCL+HGNN configuration, we apply the HyperGCL self-supervised pretraining strategy to the HGNN architecture. Subsequently, the pre-trained HGNN parameters are transferred and fine-tuned on downstream hypergraph datasets from different domains. This joint pretraining approach allows the model to integrate multi-domain structural knowledge, thereby enhancing its generalization capabilities across diverse downstream tasks.

\paragraph{Implemental Details}

We first conduct data preprocessing, including hyperedge construction, hypergraph structure optimization and vertex feature processing. When building hyperedges, any co-citation or co-director-like association forms a hyperedge; however, during dataset construction, we remove hyperedges with abnormally large degrees ($>40$, typically accounting for a small proportion) to reduce structural noise, and further eliminate isolated vertices from the remaining hypergraph to ensure valid node connectivity. For vertex features, we use the bert-base-uncased\cite{bert} model to extract 768-dimensional textual features, then compress the dimension to 256 via Singular Value Decomposition (SVD) to lower computational complexity.

In the pretraining phase, we fix HGNN layers at 2, with a hidden dimension of 128 for both HGNN and HGNN$^+$. The Vertex Knowledge Embedding module uses 4 hierarchical layers ($T=4$, validated in \cite{chien2021node}); the number of hierarchical label clusters $K$ is calculated as $K=N/B$ ($B=100$, $N$ is sample count), following \cite{zhang2021fast}—excessively small/large $K$ degrades performance, especially small $K$ leading to overcrowded clusters. \revs{While we adopt this fixed heuristic for consistency with prior work, we acknowledge that adaptive or dynamic determination of $K$ within constrained ranges could further enhance model adaptability and performance, and we leave this as an important direction for future exploration.} For "Multi-Domain Hypergraphs Sampling", we adopt BFS: neighbor vertices are randomly sampled at $80\% \sim 100\%$, expansion depth is constrained by total sampled vertices, and the sampling rate is dynamically adjusted to cover at least third-order neighbors of each vertex. After sampling 500 vertices per domain, we cluster them into 5 groups and generate hyperedges via second-order KNN. The Adam optimizer with a learning rate of 0.001 is used here.

In the fine-tuning phase, vertex classification serves as the downstream task, with a single-layer MLP as the classifier and only pre-trained HGNN layers fine-tuned to retain generalizable pretraining knowledge. For fair comparison, supervised baselines (MLP, HGNN, HGNN$^+$, HNHN, UniGCN) have a hidden dimension of 128 (matching the pretraining backbone) and use Adam with a learning rate of 0.01 (consistent with the downstream task). Data splitting follows the C-way-1-shot setting\cite{zhao2024all}: training data is built under this setting, 100 vertices per class are used as the validation set, and the rest as the test set. \revs{For datasets with temporal information (e.g., OGBN-Arxiv), all samples were shuffled across publication years before random splitting, since chronological order is not the focus of our study and random sampling avoids potential bias.} Performance is evaluated by the mean and variance of five independent runs to ensure reliability.


\begin{table*}[!ht]
\centering
\caption{Experimental results on five text-attributed hypergraph datasets.}
\label{exp:tab:main1}
\begin{threeparttable}
\begin{tabularx}{\textwidth}{ll*{5}{C}}
\toprule
 & Methods & Aminer-Text & Cora-CA-Text & Cora-CC-Text & Pubmed-CA-Text & Pubmed-CC-Text \\ 
\midrule
\multirow{5}{*}{Supervised}  
 & MLP & $0.3573_{\pm.01}$ & $0.2399_{\pm.03}$ & $0.3063_{\pm.04}$ & $0.4123_{\pm.02}$ & $0.3168_{\pm.00}$ \\
 & HGNN & $0.3916_{\pm.01}$ & $0.2571_{\pm.01}$ & $0.3570_{\pm.04}$ & $0.4463_{\pm.05}$ & $0.3248_{\pm.09}$  \\
 & HGNN$^+$ & $0.3976_{\pm.01}$ & $0.2575_{\pm.04}$ & $0.3603_{\pm.05}$ & $0.4397_{\pm.03}$ & $0.3562_{\pm.00}$  \\
 & HNHN & $0.3813_{\pm.02}$ & $0.2527_{\pm.03}$ & $0.3525_{\pm.05}$ & $0.4359_{\pm.04}$ & $0.3574_{\pm.08}$  \\
 & UniGCN & $0.3845_{\pm.01}$ & $0.2538_{\pm.04}$ & $0.3530_{\pm.05}$ & $0.4573_{\pm.05}$ & $0.3705_{\pm.06}$  \\ 
\midrule
\multirow{4}{*}{IP + Finetune} 
 & HyperGCL + HGNN & $0.3882_{\pm.01}$ & $0.2415_{\pm.05}$ & $0.3268_{\pm.08}$ & $0.3749_{\pm.01}$ & $0.3884_{\pm.01}$ \\
 & HyperGCL + HGNN$^+$ & $0.3948_{\pm.02}$ & $0.2436_{\pm.05}$ & $0.3443_{\pm.06}$ & $0.4010_{\pm.01}$ & $0.3591_{\pm.06}$ \\
 & SS-HT + HGNN & $0.3901_{\pm.02}$ & $0.2437_{\pm.03}$ & $0.3322_{\pm.07}$ & $0.4404_{\pm.02}$ & $0.3848_{\pm.00}$ \\
 & SS-HT + HGNN$^+$ & $0.3894_{\pm.03}$ & $0.2583_{\pm.04}$ & $0.3382_{\pm.04}$ & $0.4539_{\pm.03}$ & $0.3875_{\pm.02}$ \\ 
\midrule
\multirow{4}{*}{Hyper-FM + Finetune} 
 & HyperGCL + HGNN & $\mathbf{0.4325_{\pm.02}}$ & $0.2633_{\pm.03}$ & $0.3706_{\pm.04}$ & $0.4603_{\pm.04}$ & $\mathbf{0.4009_{\pm.01}}$ \\
 & HyperGCL + HGNN$^+$ & $0.4174_{\pm.02}$ & $0.2654_{\pm.03}$ & $\mathbf{0.3961_{\pm.04}}$ & $0.4641_{\pm.03}$ & $0.3972_{\pm.01}$ \\
 & SS-HT + HGNN & $0.3976_{\pm.02}$ & $\mathbf{0.2777_{\pm.03}}$ & $0.3525_{\pm.05}$ & $0.4692_{\pm.03}$ & $0.3942_{\pm.00}$ \\
 & SS-HT + HGNN$^+$ & $0.3995_{\pm.02}$ & $0.2707_{\pm.06}$ & $0.3551_{\pm.05}$ & $\mathbf{0.4708_{\pm.04}}$ & $0.3948_{\pm.00}$ \\ 
\midrule
\multicolumn{2}{c}{IMP(\%)} 
& $\mathbf{10.4\%}$ & $\mathbf{8.0\%}$ & $\mathbf{9.9\%}$ & $\mathbf{7.1\%}$ & $\mathbf{23.4\%}$  \\ 
\bottomrule
\end{tabularx}
\begin{tablenotes}
\footnotesize
\item The bold numbers indicate the best results in each column. The "IP" stands for Isolated Pretraining. 
\item The ``IMP'' denotes the performance gain of the optimal method under the ``Hyper-FM + Finetune'' setting relative to its counterpart without pretraining.
\end{tablenotes}
\end{threeparttable}
\end{table*}

\begin{table*}[!ht]
\centering
\scriptsize
\caption{Experimental results on another six text-attributed hypergraph datasets.}
\label{exp:tab:main2}
\begin{threeparttable}
\begin{tabularx}{\textwidth}{llX*{6}X}
\toprule
 & Methods & Arxiv-Text & Movielens-Text & IMDB-Text & GoodBook-Text & PPI-Text & OGBN-Arxiv \\ 
\midrule
\multirow{5}{*}{Supervised}  
 & MLP & $0.3345_{\pm .02}$ & $0.2880_{\pm .03}$ & $0.2146_{\pm .09}$ & $0.2489_{\pm .04}$ & $0.4194_{\pm .09}$ & $0.2034_{\pm .04}$ \\
 & HGNN & $0.3523_{\pm .05}$ & $0.3261_{\pm .08}$ & $0.2533_{\pm .03}$ & $0.3554_{\pm .03}$ & $0.4306_{\pm .08}$ & $0.2348_{\pm .04}$ \\
 & HGNN$^+$ & $0.3825_{\pm .08}$ & $0.3016_{\pm .09}$ & $0.2205_{\pm .09}$ & $0.3225_{\pm .01}$ & $0.5065_{\pm .08}$ & $0.2441_{\pm .03}$ \\
 & HNHN & $0.3753_{\pm .09}$ & $0.2688_{\pm .09}$ & $0.2528_{\pm .07}$ & $0.3130_{\pm .01}$ & $0.5261_{\pm .05}$ & $0.2331_{\pm .03}$ \\
 & UniGCN & $0.3571_{\pm .02}$ & $0.3269_{\pm .07}$ & $0.2565_{\pm .05}$ & $0.3607_{\pm .04}$ & $0.4387_{\pm .07}$ & $0.2559_{\pm .04}$ \\ 
\midrule
\multirow{4}{*}{IP + Finetune} 
 & HyperGCL + HGNN & $0.3540_{\pm .06}$ & $0.2384_{\pm .02}$ & $0.2536_{\pm .03}$ & $0.3278_{\pm .03}$ & $0.4339_{\pm .07}$ & $0.2657_{\pm .04}$ \\
 & HyperGCL + HGNN$^+$ & $0.3206_{\pm .04}$ & $0.2796_{\pm .04}$ & $0.2586_{\pm .04}$ & $0.3422_{\pm .02}$ & $0.5084_{\pm .06}$ & $0.2665_{\pm .04}$ \\
 & SS-HT + HGNN & $0.3505_{\pm .04}$ & $0.3185_{\pm .05}$ & $0.2486_{\pm .08}$ & $0.3210_{\pm .01}$ & $0.4355_{\pm .04}$ & $0.2414_{\pm .04}$ \\
 & SS-HT + HGNN$^+$ & $0.3783_{\pm .07}$ & $0.3174_{\pm .08}$ & $0.2466_{\pm .08}$ & $0.3166_{\pm .01}$ & $0.4742_{\pm .03}$ & $0.2596_{\pm .04}$ \\ 
\midrule
\multirow{4}{*}{Hyper-FM + Finetune}
 & HyperGCL + HGNN & $0.4189_{\pm .04}$ & $0.3464_{\pm .08}$ & $0.2676_{\pm .04}$ & $0.3588_{\pm .04}$ & $0.5371_{\pm .08}$ & $0.2764_{\pm .01}$ \\
 & HyperGCL + HGNN$^+$ & $0.4198_{\pm .07}$ & $\mathbf{0.3470_{\pm .06}}$ & $0.2672_{\pm .04}$ & $\mathbf{0.3703_{\pm .04}}$ & $\mathbf{0.5742_{\pm .06}}$ & $\mathbf{0.2789_{\pm .01}}$ \\
 & SS-HT + HGNN & $0.4044_{\pm .05}$ & $0.3305_{\pm .05}$ & $\mathbf{0.2786_{\pm .08}}$ & $0.3589_{\pm .04}$ & $0.5403_{\pm .04}$ & $0.2759_{\pm .04}$ \\
 & SS-HT + HGNN$^+$ & $\mathbf{0.4649_{\pm .05}}$ & $0.3347_{\pm .09}$ & $0.2775_{\pm .08}$ & $0.3569_{\pm .01}$ & $0.5387_{\pm .03}$ & $0.2745_{\pm .04}$ \\ 
\midrule
\multicolumn{2}{c}{IMP(\%)} 
 & $\mathbf{21.5\%}$ & $\mathbf{15.1\%}$ & $\mathbf{9.9\%}$ & $\mathbf{14.8\%}$ & $\mathbf{13.4\%}$ & $\mathbf{14.3\%}$ \\ 
\bottomrule
\end{tabularx}
\end{threeparttable}
\end{table*}

\subsection{Results and Discussions}

\cref{exp:tab:main1,exp:tab:main2} present the experimental results of our proposed Hyper-FM on eleven TAHG datasets. From these tables, we can draw the following four key observations. 

Firstly, hypergraph neural network methods such as HGNN consistently outperform the MLP baseline across all datasets. This improvement underscores the effectiveness of leveraging hypergraph-structured associations, which facilitate the transmission of relational information and enable the model to capture more comprehensive features, thereby enhancing overall performance.
Secondly, our proposed Hyper-FM method demonstrates a substantial performance enhancement of approximately 13.4\% across all eleven datasets, with an exceptional improvement of 23.4\% observed on the Pubmed-CC-Text dataset. Notably, we also observe a significant improvement of 14.3\% on the large-scale and complex vertex classification dataset OGBN-Arxiv, which demonstrates the effectiveness of Hyper-FM in handling large-scale and complex hypergraph data. This significant uplift highlights the efficacy of the Hypergraph Foundation Model. By conducting pretraining on multi-domain hypergraph datasets, Hyper-FM successfully integrates diverse domain-specific knowledge into the hypergraph, which in turn markedly boosts the performance of target domain models during downstream tasks.
Thirdly, when comparing with pretraining-based methods, the IP+Finetune approach exhibits noticeably inferior performance relative to Hyper-FM+Finetune. In particular, the Cora-CC-Text dataset experiences negative transfer, resulting in decreased performance in the target domain.  This decline can be attributed to the sequential nature of IP+Finetune's pretraining strategy, which despite utilizing self-supervised training to acquire multi-modal hypergraph domain knowledge, is susceptible to knowledge forgetting. Additionally, the significant disparities in knowledge distributions across different domains hinder the convergence of sequential pretraining, thereby adversely affecting performance on downstream tasks.

Lastly, within the Hypergraph Foundation Model framework, the choice of self-supervised strategy, whether HyperGCL or SS-HT, has a relatively minor impact on the overall results. In contrast, the utilization of multi-domain hierarchical hypergraphs exerts a pronounced effect on performance. A similar trend is observed in the graph foundation model GCOPE\cite{zhao2024all}, reinforcing the conclusion that the construction of robust association structures and the effective integration of cross-domain association information are critical focal points for advancing hypergraph foundation models.
These observations collectively demonstrate that our Hyper-FM not only leverages the inherent advantages of hypergraph structures but also effectively integrates multi-domain knowledge through sophisticated pretraining strategies. This dual capability underpins the model's superior performance and highlights the importance of comprehensive structural and semantic knowledge integration in hypergraph-based learning frameworks.

\subsection{More Evaluations and Analysis}

In this subsection, we conduct ablation studies to evaluate the individual contributions of the \textit{Vertex Knowledge Embedding} module and the \textit{Structural Knowledge Extraction} module within our Hyper-FM on eleven TAHG datasets.

\subsubsection{Ablation Study on Vertex Knowledge Embedding}
\cref{tab:exp:v_embedding} presents the experimental results for various vertex feature extraction methods. Specifically, we compare the performance of Bag-of-Words (BoW), direct feature extraction using the LM, and our proposed approach, LM with Neighborhood Prediction fine-tuning (LM+NP). The BoW method serves as a naive baseline by representing vertex features based on word frequency, while the LM approach leverages pre-trained language models to extract semantic features directly from the textual data. The LM+NP method enhances the LM by incorporating neighborhood information through a Neighborhood Prediction task during fine-tuning.  From the results, it is evident that there is no significant difference in performance between the LM and BoW methods overall. Notably, the LM features demonstrate superior performance on the Pubmed-CA-Text, Movielens-Text, Aminer-Text, Cora-CA-Text, and Cora-CC-Text datasets, whereas the BoW features yield better results on the GoodBook-Text and PPI-Text datasets. This observation indicates that the BoW approach, despite its simplicity, is capable of capturing essential vertex feature information in certain contexts. However, when employing the proposed LM+NP method, we observe a substantial improvement in performance across all eleven datasets. Specifically, the introduction of neighborhood knowledge through fine-tuning enables the model to generate more informative and rich vertex embeddings, leading to enhanced accuracy in downstream tasks. These findings underscore the advantage of integrating neighborhood information into vertex feature representations. While both BoW and LM methods have their respective strengths, the LM+NP approach effectively combines the semantic richness of language models with neighborhood insights, resulting in more robust and accurate vertex embeddings. Consequently, the enhanced vertex knowledge embedded through our proposed method plays a crucial role in improving the overall performance of the Hypergraph Foundation Model on diverse Text-Attributed Hypergraph datasets.

\begin{table}[!ht]
\caption{Ablation study on the vertex knowledge extraction.}
\label{tab:exp:v_embedding}
\centering
\begin{tabularx}{\linewidth}{lCCC}
\toprule
\multicolumn{1}{l}{} & BoW & LM & LM+NP \\ 
\midrule
Pubmed-CA-Text & $0.4260_{\pm .03}$ & $0.4497_{\pm .01}$ & $\mathbf{0.4603_{\pm .04}}$ \\
Movielens-Text & $0.2602_{\pm .02}$ & $0.2852_{\pm .04}$ & $\mathbf{0.3464_{\pm .08}}$ \\
GoodBook-Text & $0.3297_{\pm .04}$ & $0.3142_{\pm .02}$ & $\mathbf{0.3588_{\pm .04}}$ \\
PPI-Text & $0.5032_{\pm .07}$ & $0.4839_{\pm .09}$ & $\mathbf{0.5371_{\pm .08}}$ \\
Aminer-Text & $0.3940_{\pm .01}$ & $0.4073_{\pm .01}$ & $\mathbf{0.4325_{\pm .02}}$ \\
Cora-CA-Text & $0.2101_{\pm .02}$ & $0.2526_{\pm .05}$ & $\mathbf{0.2633_{\pm .03}}$ \\
Cora-CC-Text & $0.3059_{\pm .04}$ & $0.3368_{\pm .04}$ & $\mathbf{0.3706_{\pm .04}}$ \\
Pubmed-CC-Text & $0.3977_{\pm .01}$ & $0.3967_{\pm .01}$ & $\mathbf{0.4009_{\pm .01}}$ \\
Arxiv-Text & $0.2786_{\pm .03}$ & $0.3231_{\pm .07}$ & $\mathbf{0.4189_{\pm .04}}$ \\
IMDB-Text & $0.2625_{\pm .05}$ & $0.2637_{\pm .04}$ & $\mathbf{0.2676_{\pm .04}}$ \\
OGBN-Arxiv & $0.2667_{\pm .02}$ & $0.2608_{\pm .03}$ & $\mathbf{0.2789_{\pm .01}}$ \\
\bottomrule
\end{tabularx}
\end{table}

\subsubsection{Ablation Study on Hypergraph Sampling Strategy}
The first step in constructing Hierarchical Multi-Hypergraphs involves structural sampling from multiple domain-specific hypergraph datasets. In this ablation study, we evaluate the impact of different sampling strategies on the performance of our proposed Hypergraph Foundation Model. Specifically, we compare our proposed BFS-based diffusion sampling strategy with a straightforward random vertex sampling strategy.
The random vertex sampling strategy involves selecting vertices uniformly at random from the hypergraph and retaining the hyperedges that connect these selected vertices to form a sub-hypergraph. In contrast, the BFS-based sampling strategy systematically explores the hypergraph by expanding from an initial set of vertices, thereby preserving more of the inherent structural relationships within the hypergraph.
The experimental results, as shown in \cref{exp:tab:sampling}, indicate that the BFS-based sampling strategy significantly outperforms the random sampling strategy across all evaluated datasets. This superior performance can be attributed to the ability of BFS to maintain a greater extent of the hypergraph's associative structures compared to random sampling. Additionally, as the number of training epochs increases, the BFS strategy more effectively reconstructs the original hypergraph structural patterns, thereby minimizing the loss of structural information. Consequently, the BFS-based approach achieves a better balance between training speed and accuracy, demonstrating its efficacy in preserving essential hypergraph structures during the sampling process. These findings highlight the importance of the sampling strategy in the construction of Hierarchical Multi-Hypergraphs. By retaining more meaningful structural information, the BFS-based sampling approach enhances the model's ability to learn robust and generalizable representations, ultimately leading to improved performance on downstream tasks.

\begin{table}[!ht]
\caption{Ablation study on the sampling strategy of sub-hypergraphs.}
\label{exp:tab:sampling}
\centering
\begin{tabularx}{\linewidth}{XCC}
\toprule
 & Random Sampling & BFS Sampling \\ 
\midrule
Pubmed-CA-Text & $0.4340_{\pm .01}$ & $\mathbf{0.4603_{\pm .04}}$ \\
Movielens-Text & $0.3168_{\pm .09}$ & $\mathbf{0.3464_{\pm .08}}$ \\
GoodBook-Text & $0.3351_{\pm .02}$ & $\mathbf{0.3588_{\pm .04}}$ \\
PPI-Text & $0.5048_{\pm .09}$ & $\mathbf{0.5371_{\pm .08}}$ \\
Aminer-Text & $0.4217_{\pm .02}$ & $\mathbf{0.4325_{\pm .02}}$ \\
Cora-CA-Text & $0.2462_{\pm .02}$ & $\mathbf{0.2633_{\pm .03}}$ \\
Cora-CC-Text & $0.3224_{\pm .05}$ & $\mathbf{0.3706_{\pm .04}}$ \\
Pubmed-CC-Text & $0.3902_{\pm .01}$ & $\mathbf{0.4009_{\pm .01}}$ \\
Arxiv-Text & $0.3960_{\pm .05}$ & $\mathbf{0.4189_{\pm .04}}$ \\
IMDB-Text & $0.2661_{\pm .04}$ & $\mathbf{0.2676_{\pm .04}}$ \\
OGBN-Arxiv & $0.2655_{\pm .02}$ & $\mathbf{0.2789_{\pm .01}}$ \\
\bottomrule
\end{tabularx}
\end{table}

\begin{figure*}[!ht]
    \centering
    \includegraphics[width=\linewidth]{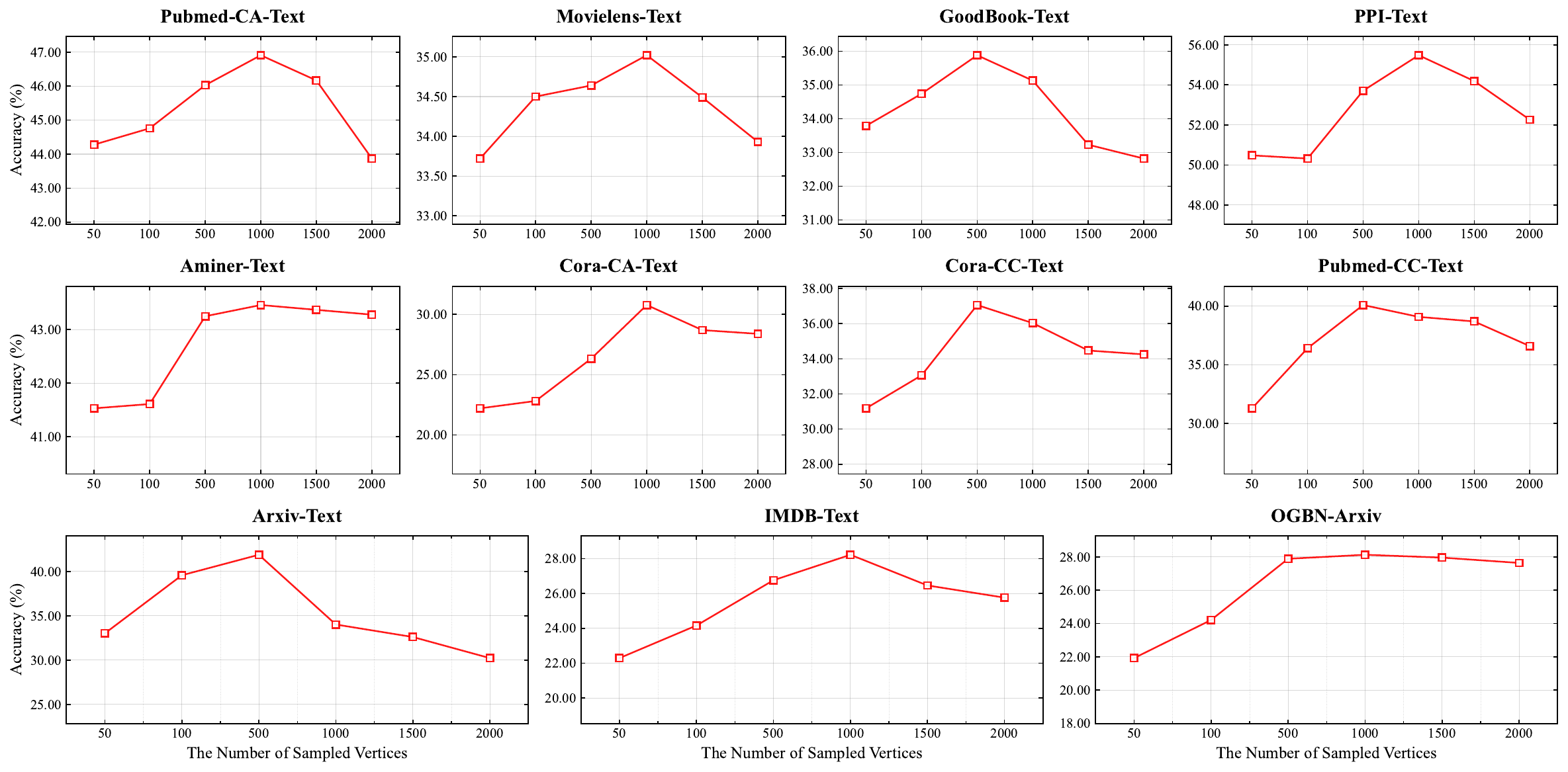}
    \caption{Experimental results of ablation on the vertex number for the sub-hypergraph sampling for each hypergraph domain dataset.}
    \label{fig:num_sampling}
\end{figure*}

Furthermore, we conduct an ablation study to investigate the impact of the number of sampled vertices in the domain hypergraph sampling process. The experimental results are illustrated in \cref{fig:num_sampling}. From the figure, it is evident that the performance across the datasets achieves its highest point when the number of sampled vertices is approximately 1000. When the number of sampled vertices is reduced to 500, the performance remains nearly unchanged. Considering both computational complexity and performance efficiency, we set the default vertex sampling count to 500. Additionally, our analysis reveals that sampling an excessively low or high number of vertices leads to a decline in performance. Specifically, sampling too few vertices results in a substantial loss of hypergraph structural knowledge, thereby degrading the model's performance. On the other hand, sampling too many vertices introduces increased structural redundancy within each domain. This redundancy does not introduce more domain structural patterns, thus restricting the effective transmission of information between domains and hampers the learning of general structural knowledge, ultimately diminishing the utility of the pre-trained model. These findings highlight the importance of selecting an optimal number of sampled vertices to balance the preservation of structural information and the efficiency of knowledge transfer across domains. By setting the vertex sampling count to 500, we achieve a favorable trade-off between maintaining essential hypergraph structures and ensuring computational feasibility, thereby enhancing the overall performance of the Hypergraph Foundation Model.

\subsubsection{Ablation Study on Structure Knowledge Extraction}
In our Hyper-FM, the core of structural knowledge extraction lies in the construction of hierarchical multi-hypergraphs. The main experiments have already demonstrated that jointly integrating multi-domain hypergraphs can significantly enhance the model's ability to extract structural knowledge from hypergraphs. In this ablation study, we further investigate the impact of the connection structure within the jointly integrated multi-domain hypergraphs. Specifically, we evaluate two different configurations: (1) without using clustering, directly connecting vertices from corresponding domain-specific hypergraphs (resulting in a non-hierarchical hypergraph), (2) using clustering to construct hierarchical hypergraphs and varying the number of clusters to observe its effect on the final performance. The experimental results are presented in \cref{exp:tab:clustering}. When constructing multi-domain hypergraphs without clustering—indicated by a clustering number of 1 in the table—versus using clustering to build hierarchical hypergraphs, we can make two key observations. 

Firstly, employing clustering to construct hierarchical hypergraphs leads to a significant performance improvement. This enhancement is attributable to the inherent complexity of hypergraph structures and the natural formation of distinct semantic clusters within the same domain due to varying vertex labels. Directly connecting multi-domain datasets that contain different semantic meanings without clustering impedes the effective transmission of information across the hypergraph, resulting in a decline in performance. Secondly, we observe that the number of clusters used in the clustering process has a varying impact on the capability of the Hypergraph Foundation Model. The optimal number of clusters in the hierarchical hypergraph is found to be approximately equal to the number of classes in the dataset. This observation indirectly suggests that clustering based on vertex semantic information effectively reflects the distribution of vertex labels. Consequently, the setting of cluster numbers can be guided by the number of classes present in the original data, providing a practical heuristic for setting this hyperparameter. These findings underscore the critical role of clustering in constructing hierarchical multi-hypergraphs, as it facilitates the preservation and effective utilization of semantic and structural information within and across domains. By aligning the number of clusters with the underlying class structure of the data, the model can better capture and leverage the nuanced relationships inherent in hypergraph-structured datasets, leading to enhanced performance in downstream tasks.

\begin{table*}[!ht]
\centering
\caption{Ablation study on varying the number of clusters in the hierarchical multi-hypergraph.}
\label{exp:tab:clustering}
\begin{tabular}{lccccccc}
\toprule
\multicolumn{1}{c}{\multirow{3}{*}{}} & \multicolumn{6}{c}{Number of Clustering} & Number of  \\ 
\multicolumn{1}{l}{} & 1 & 2 & 3 & 4 & 5 & 6 & Classes \\  
\midrule
Pubmed-CA-Text & $0.4332_{\pm .03}$ & $0.4460_{\pm .03}$ & $0.4522_{\pm .05}$ & $0.4465_{\pm .01}$ & $\mathbf{0.4603_{\pm .04}}$ & $0.4432_{\pm .04}$ & 3 \\
Movielens-Text & $0.2884_{\pm .02}$ & $0.3132_{\pm .09}$ & $0.3281_{\pm .07}$ & $0.3304_{\pm .05}$ & $\mathbf{0.3464_{\pm .08}}$ & $0.3217_{\pm .07}$ & 5 \\
GoodBook-Text & $0.3378_{\pm .06}$ & $0.3307_{\pm .03}$ & $0.3294_{\pm .03}$ & $0.3478_{\pm .05}$ & $\mathbf{0.3588_{\pm .04}}$ & $0.3227_{\pm .04}$ & 4 \\
PPI-Text & $0.4919_{\pm .09}$ & $0.5371_{\pm .09}$ & $0.5581_{\pm .09}$ & $\mathbf{0.5613_{\pm .09}}$ & $0.5371_{\pm .08}$ & $0.5016_{\pm .06}$ & 4 \\
Aminer-Text & $0.4174_{\pm .02}$ & $0.4195_{\pm .02}$ & $0.4201_{\pm .02}$ & $0.4284_{\pm .02}$ & $\mathbf{0.4325_{\pm .02}}$ & $0.4278_{\pm .01}$ & 3 \\
Cora-CA-Text & $0.2535_{\pm .05}$ & $0.2645_{\pm .04}$ & $0.2637_{\pm .04}$ & $0.2716_{\pm .04}$ & $0.2633_{\pm .03}$ & $\mathbf{0.2685_{\pm .03}}$ & 7 \\
Cora-CC-Text & $0.3174_{\pm .03}$ & $0.3431_{\pm .04}$ & $0.3551_{\pm .04}$ & $0.3595_{\pm .05}$ & $0.3706_{\pm .04}$ & $\mathbf{0.3735_{\pm .04}}$ & 7 \\
Pubmed-CC-Text & $0.3531_{\pm .06}$ & $0.3577_{\pm .06}$ & $0.3856_{\pm .01}$ & $0.3918_{\pm .01}$ & $\mathbf{0.4009_{\pm .01}}$ & $0.3948_{\pm .01}$ & 3 \\
Arxiv-Text & $0.3345_{\pm .05}$ & $0.3315_{\pm .05}$ & $0.3198_{\pm .04}$ & $0.3329_{\pm .08}$ & $\mathbf{0.4189_{\pm .04}}$ & $0.3270_{\pm .08}$ & 5 \\
IMDB-Text & $0.2417_{\pm .01}$ & $0.2452_{\pm .01}$ & $0.2520_{\pm .04}$ & $0.2820_{\pm .05}$ & $0.2676_{\pm .04}$ & $\mathbf{0.2748_{\pm .04}}$ & 5 \\
OGBN-Arxiv & $0.2337_{\pm .04}$ & $0.2401_{\pm .04}$ & $0.2637_{\pm .04}$ & $0.2728_{\pm .02}$ & $\mathbf{0.2789_{\pm .01}}$ & $0.2756_{\pm .04}$ & 40 \\
\bottomrule
\end{tabular}
\end{table*}

\begin{figure*}[!ht]
    \centering
    \includegraphics[width=\linewidth]{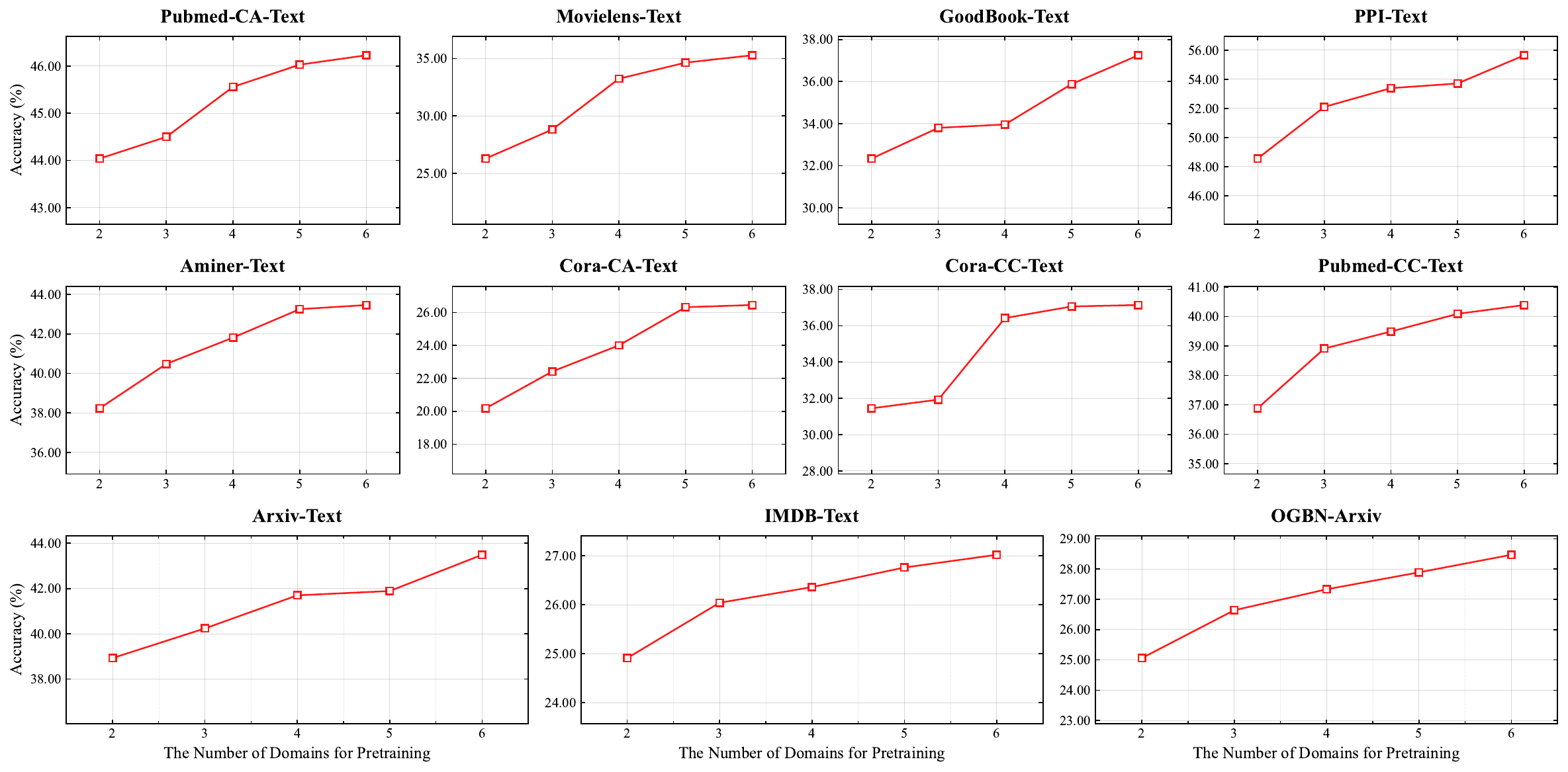}
    \caption{Experimental results of ablation on the number of domains for pretraining.}
    \label{fig:more_domain}
\end{figure*}

\subsection{Scaling Law of Hypergraph Foundation Model}
In most foundation models, the model's capability increases with the amount of data, as observed in CV\cite{dido} and NLP models\cite{gpt3}. However, relational data differ in their representation from these domains. Unlike CV data, which can be directly visualized, or NLP data, which can be directly interpreted, relational data primarily model the connections within real-world networks. As illustrated in \cref{fig:num_sampling}, the performance of foundation models does not necessarily improve with an increase in the volume of relational data. In this subsection, we explore the scaling law of the Hypergraph Foundation Model from a different perspective—the number of domains. The experimental results are presented in \cref{fig:more_domain}. Each domain-specific hypergraph dataset possesses unique connection patterns, and an increase in the number of domains introduces a variety of relational data patterns. This diversity injects more comprehensive knowledge into the Hypergraph Foundation Model. Remarkably, the experimental findings reveal that as the number of domains increases, the performance of the Hypergraph Foundation Model on downstream tasks continues to improve. This trend suggests that incorporating a greater variety of relational structures, each with its distinct patterns, enhances the model's ability to learn and generalize. \textbf{The addition of diverse domain-specific knowledge contributes more effectively to the foundation model's power than merely increasing the number of vertices and hyperedges within the hypergraphs.} These results highlight that the richness and diversity of relational structures across multiple domains play a crucial role in scaling the capabilities of hypergraph-based foundation models. By leveraging information from varied connection patterns, the Hypergraph Foundation Model can better capture complex relationships and dependencies, leading to superior performance in downstream applications. This finding underscores the importance of multi-domain integration in the development and scaling of foundation models for relational data.

\section{Conclusion}
In this work, we introduce Hyper-FM, the first Hypergraph Foundation Model for multi-domain knowledge extraction. By implementing Hierarchical High-Order Neighbor Guided Vertex Knowledge Embedding, we integrate domain-specific information into vertex features, addressing varying dimensions and structural complexities. Our Hierarchical Multi-Hypergraph Guided Structural Knowledge Extraction method allows scalable and efficient extraction of structural knowledge from diverse datasets, mitigating negative transfer and domain-specific blending. Additionally, we curate 11 text-attributed hypergraph (TAHG) datasets to bridge hypergraph neural networks and large language models. Empirical evaluations show that Hyper-FM outperforms baseline methods by approximately 10\% across these datasets, demonstrating its robustness and effectiveness. We also establish a novel scaling law, revealing that increasing domain diversity significantly enhances model performance, unlike merely expanding vertex and hyperedge counts. Hyper-FM sets a new benchmark for hypergraph foundation models, providing a scalable and effective solution for multi-domain knowledge extraction. Future work will explore expanding TAHG datasets and applying Hyper-FM to a broader range of domains to further enhance its versatility and impact.

\bibliographystyle{IEEEtran}
\bibliography{IEEEabrv,mybib}

\begin{IEEEbiography}[{\vspace{-1cm} 
\includegraphics[width=1in,height=1.25in,clip,keepaspectratio]{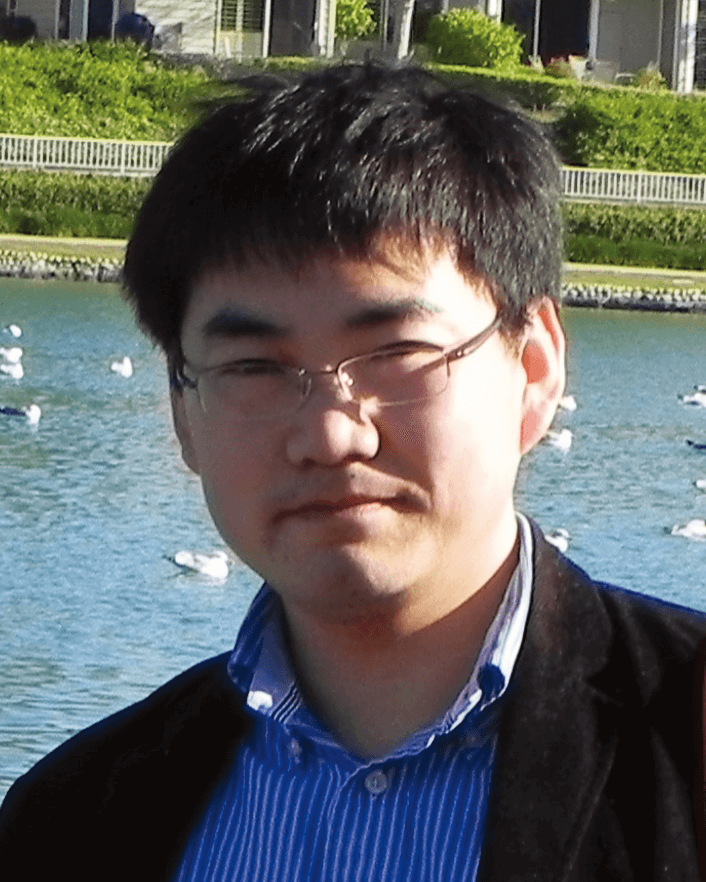}}]{Yue Gao}
is an associate professor with the School of Software, Tsinghua University. He received the B.S. degree from the Harbin Institute of Technology, Harbin, China, and the M.E. and Ph.D. degrees from Tsinghua University, Beijing, China.
\end{IEEEbiography}

\begin{IEEEbiography}[{\vspace{-1cm}
\includegraphics[width=1in,height=1.25in,clip,keepaspectratio]{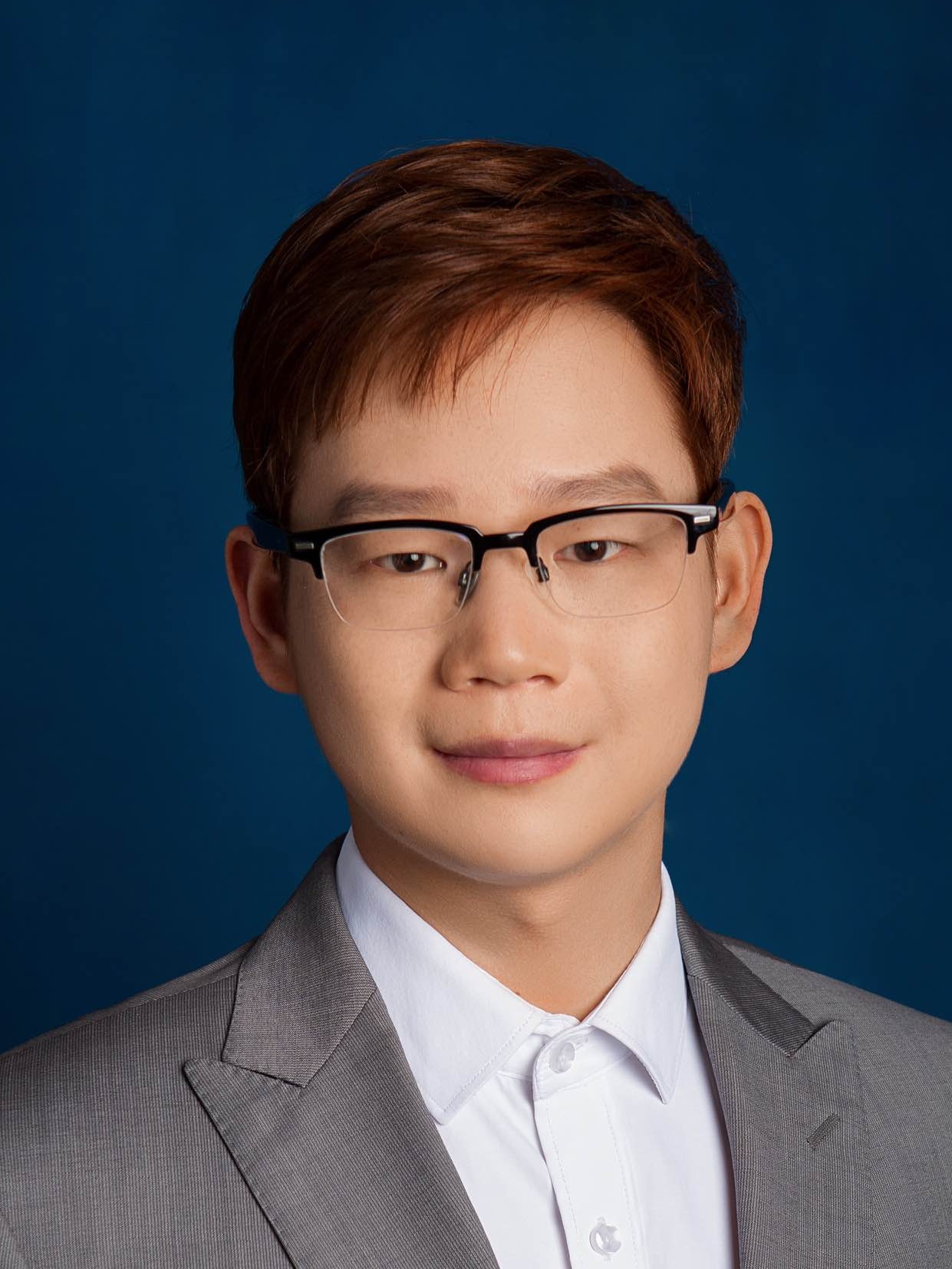}}]{Yifan Feng} 
received the BE degree in computer science and technology from Xidian University, Xi'an, China, in 2018, the MS degree from Xiamen University, Xiamen, China, in 2021, and the Ph.D degree from Tsinghua University, Beijing, China, in 2025. His research interests include hypergraph learning, multi-modal learning, and large language model.
\end{IEEEbiography}
\begin{IEEEbiography}[{\vspace{-1cm}
\includegraphics[width=1in,height=1.25in,clip,keepaspectratio]{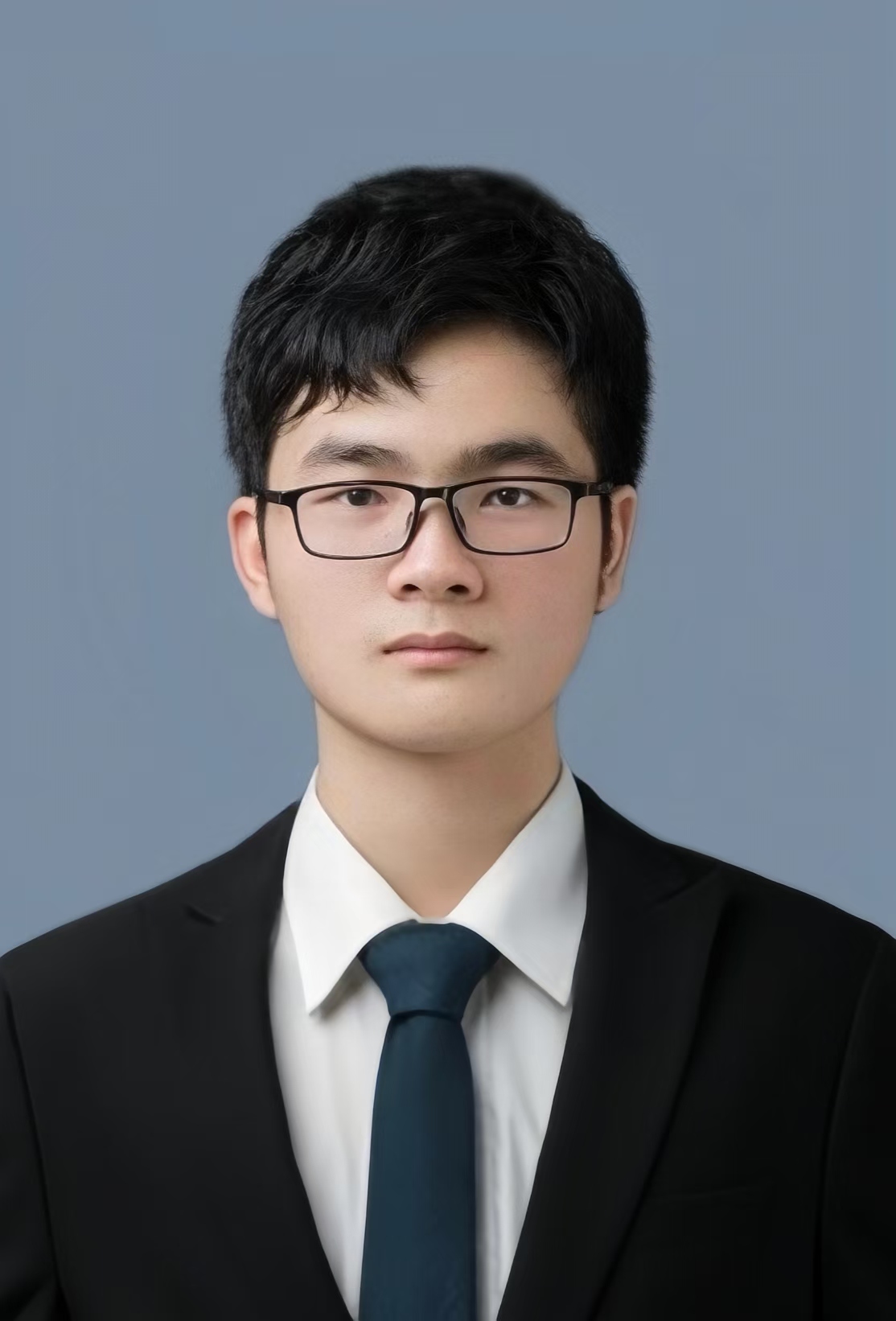}}]{Shiquan Liu} 
received the BE degree from Xi'an Jiaotong University, Xi'an, China, in 2023. He is currently working toward the master's degree in artificial intelligence from Xi'an Jiaotong University.
\end{IEEEbiography}

\begin{IEEEbiography}[{\vspace{-1cm}
\includegraphics[width=1in,height=1.25in,clip,keepaspectratio]{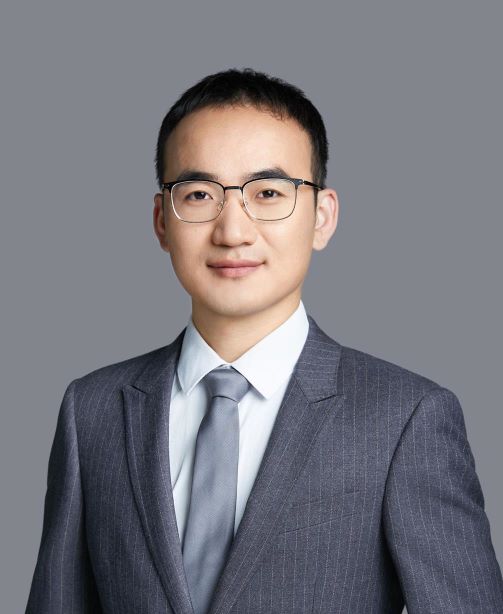}}]{Xiangmin Han} 
is a postdoctoral at the School of Software, Tsinghua University. He received his B.S. and M.E. degrees from Beijing Union University, Beijing, China, and his Ph.D. degrees from Shanghai University, Shanghai, China. His research interests include medical image analysis, Brain network analysis, and hypergraph computation.
\end{IEEEbiography}

\vfill

\begin{IEEEbiography}[{
\includegraphics[width=1in,height=1.25in,clip,keepaspectratio]{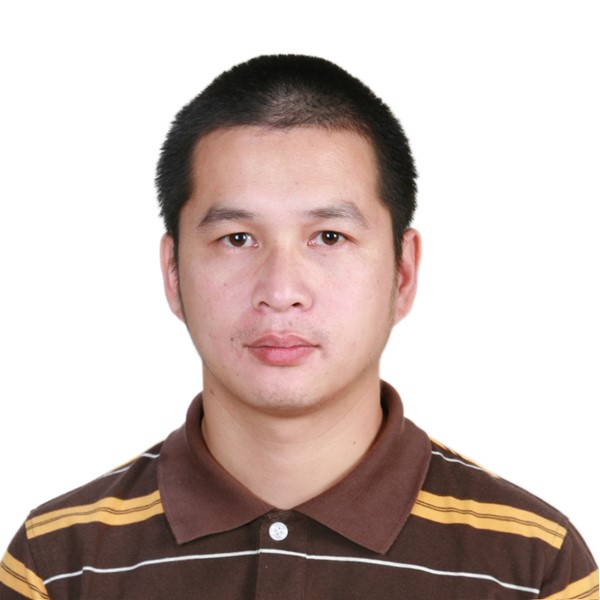}}]{Shaoyi Du} 
received double Bachelor degrees in computational mathematics and in computer science in 2002 and received his M.S. degree in applied mathematics in 2005 and Ph.D. degree in pattern recognition and intelligence system from Xi’an Jiaotong University, China in 2009. He is a professor at Xi’an Jiaotong University. His research interests include computer vision, machine learning and pattern recognition.
\end{IEEEbiography}

\begin{IEEEbiography}[{
\includegraphics[width=1in,height=1.25in,clip,keepaspectratio]{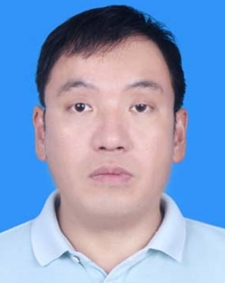}}]{Zongze Wu} 
(Member, IEEE) received the B.S. degree in material forming and control, the M.S. degree in control science and engineering, and the Ph.D. degree in pattern reorganization and intelligence system from Xi’an Jiaotong University, Xi’an, China, in 1999, 2002, and 2005, respectively. He is currently a Professor with the College of Mechatronics and Control Engineering, Shenzhen University, China, where he is also with the Guangdong Laboratory of Artificial Intelligence and Digital Economy (SZ). He is the author and coauthor of 70 SCI indexed papers, which include IEEE TRANSACTIONS ON AUTOMATIC CONTROL, IEEE TRANSACTIONS ON CYBERNETICS, IEEE TRANSACTIONS ON SYSTEM, MAN, AND CYBERNETICS: SYSTEMS, IEEE/CAAJOURNAL OF AUTOMATICASINICA, IEEE TRANSACTIONS ON IMAGE PROCESSING, Pattern Recognition, and Signal Processing.
\end{IEEEbiography}

\begin{IEEEbiography}[{\vspace{-1cm} 
\includegraphics[width=1in,height=1.25in,clip,keepaspectratio]{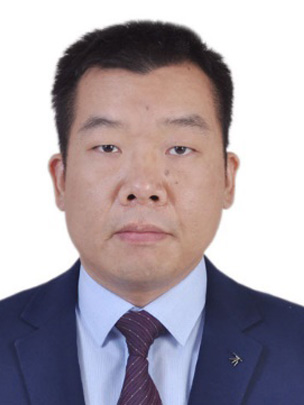}}]{Han Hu}
received the B.E. and Ph.D. degrees from the University of Science and Technology of China, China, in 2007 and 2012, respectively. He is currently a Professor with the School of Information and Electronics, Beijing Institute of Technology, China. His research interests include multimedia networking, edge intelligence, and space-air-ground integrated network. He received several academic awards, including the Best Paper Award of the IEEE TCSVT 2019, the Best Paper Award of the IEEE Multimedia Magazine 2015, and the Best Paper Award of the IEEE Globecom 2013. He has served as an Associate Editor of IEEE TMM and Ad Hoc Networks, and a TPC member of Infocom, ACM MM, AAAI, and IJCAI.
\end{IEEEbiography}

\vfill

\end{document}